\documentclass[runningheads]{llncs}

\usepackage[T1]{fontenc}
\usepackage[pdftex]{graphicx}
\usepackage{verbatim}
\usepackage{cite}
\usepackage{amssymb}
\usepackage{amsmath}
\usepackage{mathbbol} % for \mathbb{1}
\usepackage{float}
\usepackage{mathtools}
\usepackage[linesnumbered, algoruled, vlined]{algorithm2e}
\usepackage{caption}
\usepackage{subcaption}
\usepackage{tikz}
\usepackage{hyperref}
\usepackage{wrapfig}
\usepackage{float}
\usepackage{mathrsfs}
\usepackage{array}
\usepackage[export]{adjustbox}
\usepackage{cleveref}
\usepackage{macros}
\usepackage{algpseudocode}
\usepackage{placeins}
\usepackage[colorinlistoftodos]{todonotes}
\usetikzlibrary{arrows, shapes, automata, positioning}
\tikzset{auto, >=stealth}
\tikzset{every edge/.append style={shorten >= 1pt}}
\tikzset{main node/.style={circle,draw,minimum size=1cm,inner sep=0pt},
}

\newcommand{\zhe}[1]{[\textcolor{red}{\textbf{ZX:} #1}]}

\renewcommand\zhe[1]{}
\renewcommand\todo[2][]{}

\allowdisplaybreaks
\begin{document}\sloppy

\title{Reinforcement Learning with Temporal-Logic-Based Causal Diagrams }
% \footnote{The first three authors contribute equally to this paper.}
\author{Yash Paliwal\inst{\thanks{The first three authors contributed equally.}1} \and
Rajarshi Roy\inst{\ast2} \and Jean-Rapha\"el Gaglione\inst{\ast3}  \and
Nasim Baharisangari\inst{1} \and
Daniel Neider\inst{4,5} \and
Xiaoming Duan\inst{\thanks{The published version of the paper \cite{Paliwa23} contained an oversight.
Duan performed the work while affiliated with the University of Texas at Austin. Present address: Department of Automation, Shanghai Jiao Tong University.}3} \and
Ufuk Topcu\inst{3} \and Zhe Xu\inst{1} }
\authorrunning{F. Author et al.}
\institute{Arizona State University, Arizona, USA \and
	Max Planck Institute for Software Systems, Kaiserslautern, Germany \and
 University of Texas at Austin, Texas, USA \and 
	TU Dortmund University, Dortmund, Germany \and
	Center for Trustworthy Data Science and Security, University Alliance Ruhr, Germany 
 }
%  \and Department of Automation, Shanghai Jiao Tong University
\maketitle              % typeset the header of the contribution
\begin{abstract}
We study a class of reinforcement learning (RL) tasks where the objective of the agent is to accomplish temporally extended goals. In this setting, a common approach is to represent the tasks as deterministic finite automata (DFA) and integrate them into the state-space for RL algorithms. However, while these machines model the reward function, they often overlook the causal knowledge about the environment. To address this limitation, we propose the Temporal-Logic-based Causal Diagram (TL-CD) in RL, which captures the temporal causal relationships between different properties of the environment. We exploit the TL-CD to devise an RL algorithm in which an agent requires significantly less exploration of the environment. To this end, based on a TL-CD and a task DFA, we identify configurations where the agent can determine the expected rewards early during an exploration. Through a series of case studies, we demonstrate the benefits of using TL-CDs, particularly the faster convergence of the algorithm to an optimal policy due to reduced exploration of the environment.

% We prove that this algorithm can learn the optimal policy in the limit. A common approach in this setting is to model the goals as finite state machines and these machines are, thereafter, integrated in the state-space to perform RL algorithms. While the finite state machines simply model the reward function, they often overlook the known causal knowledge about the underlying environment. Thus, in this paper, we introduce the Temporal-Logic-based Causal Diagram (TL-CD) in RL to capture the temporal causal relations between various properties of the environment. We exploit TL-CD to devise an RL algorithm in which an agent requires significantly less exploration of the environment. To this end, based on a TL-CD and a finite state reward machine, we identify configurations where the agent can determine the expected rewards early during an exploration. We prove that such an algorithm can indeed identify the optimal policy in the limit. In our evaluation, we demonstrate the benefits of using TL-CDs in a number of case-studies. In particular, we show that by using a TL-CD, the algorithm converges faster to an optimal policy due to less exploration of the environment.

% \todo{
% TL causal diagram -> TL formula -> Causal DFA
% Task DFA
% motivation of detecting when early stopping "expedite RL" "predict a reward"
% "using causal knowledge, identify"
% training efficiency in number of cumulated time steps.
% }

\end{abstract}

\section{Introduction}
In many reinforcement learning (RL) tasks, the objective of the agent is to accomplish temporally extended goals that require multiple actions to achieve.
One common approach to modeling these goals is to use finite state machines.
However, these machines only model the reward function and do not take into account the causal knowledge of the underlying environment, which can limit the effectiveness of the RL algorithms \cite{Aksaray2016,Lixiao2017,Fu2014ProbablyAC,Min2017,Alshiekh2018SafeRL,xu2019joint,AFRAI2021,neider2021advice,zhe_ijcai2019}.

 Moreover, online RL, including in the non-Markovian setting, often requires extensive interactions with the environment.
 This impedes the adoption of RL algorithms in real-world applications due to the impracticality of expensive and/or unsafe data collection during the exploration phase.
 %  Reinforcement learning (RL) is a subfield of artificial intelligence that focuses on training agents to interact with an environment and learn through trial-and-error to achieve a desired goal. Despite the fact that many causal inference and causal discovery approaches have been proposed in the past few decades, their application to reinforcement learning (RL) still faces several fundamental challenges.

To address these limitations, in this paper we propose Temporal-Logic-based Causal Diagrams (TL-CDs) which can capture the temporal causal relationships between different properties of the environment, allowing the agent to make more informed decisions and require less exploration of the environment.
TL-CDs combine temporal logic, which allows for reasoning about events over time, with causal diagrams, which represent the causal relationships between variables.
By using TL-CDs, the RL algorithm can exploit the causal knowledge of the environment to identify configurations where the agent can determine the expected rewards early during an exploration, leading to faster convergence to an optimal policy.

We introduce an RL algorithm that leverages TL-CDs to achieve temporally extended goals.
We show that our algorithm requires significantly less exploration of the environment than traditional RL algorithms that use finite state machines to model goals.
By using TL-CDs, our algorithm identifies configurations where the agent can determine the expected rewards early during exploration, reducing the number of steps required to achieve the goal.

% We prove that our algorithm can identify the optimal policy in the limit and demonstrate its effectiveness through a series of case studies.

% In this paper, we introduce Temporal Logic based Causal Diagram (TL-CD) in RL to capture the temporal causal relations between various properties of the environment. We exploit TL-CD to devise an RL algorithm in which an agent requires significantly less exploration of the environment.
% In this paper, we propose RL with causal reward machines (CRMs) where each CRM consists of two parts: (1) a reward machine which captures the temporal or dynamic aspect of the reward structure and (2) a causal diagram (about the events on the transitions of the reward machine) which aims to reflect the underlying Structural Causal Model (SCM) of the events in the environment. This approach is more flexible and valuable than conventional methods, since the learned CRM can capture temporal causal relationships among the events, instead of just association rules. The policy evaluation and improvement can become much more efficient as the accumulated rewards of the entire trajectory can be predicted using the causal relationships between earlier and later events.

\section{Motivating Example}\label{sec:motivating-example}

\begin{figure}[ht] % fig:example-seed
    \centering
    \begin{subfigure}[b]{.40\linewidth}
        \centering
        \begin{tikzpicture}[
    scale=.7,
    thick,
    every node/.append style={transform shape},
]
% == NODES =================================================================== %
\node[state,initial] (uInit)
    at (0,0)
    {$\AutState[task]_{0}$};
\node[state] (uA)
    at (2,0.5)
    {$\AutState[task]_{1}$};
\node[state] (uB)
    at (2,-0.5)
    {$\AutState[task]_{2}$};
\node[state,accepting] (uEnd)
    at (4,0)
    {$\AutState[task]_{3}$};

% == EDGES =================================================================== %
\path[->,sloped]

(uInit) % -------------------------------------------------------------------- %
edge[loop above] node[sloped=false]
    {$\lnot p \land \lnot s$}
    (u0)
edge[] node[]
    {$p \land \lnot s$}
    (uA)
edge[] node[swap]
    {$s$}
    (uB)

(uA) % ----------------------------------------------------------------------- %
edge[loop above] node[sloped=false]
    {$\lnot g$}
    (uA)
edge[] node[]
    {$g$}
    (uEnd)

(uB) % ----------------------------------------------------------------------- %
edge[loop below] node[sloped=false]
    {$\lnot b$}
    (uB)
edge[] node[swap]
    {$b$}
    (uEnd)

(uEnd) % --------------------------------------------------------------------- %
edge[loop right] node[sloped=false]
    {$\ltrue$}
    (uEnd)

;

\end{tikzpicture}
        \caption{Task DFA $\Autom[task]$}
        \label{fig:example-seed:TaskDFA}
    \end{subfigure}%
    \begin{subfigure}[b]{.30\linewidth}
        \centering
        \begin{tikzpicture}[
    scale=.7,
    thick,
    every node/.append style={transform shape},
]
% == NODES =================================================================== %
\node[event] (psi1)
    at (0,-0)
    {$p$};
\node[event] (phi1)
    at (0,-1.5)
    {$\lX g$};
    
\node[event] (psi2)
    at (2,-0)
    {$s$};
\node[event] (phi2)
    at (2,-1.5)
    {$\lG \lnot \lnext b$};

% == EDGES =================================================================== %
\path[->,sloped]
(psi1) edge[cause] node {} (phi1)
(psi2) edge[cause] node {} (phi2)
;

\end{tikzpicture}
        \caption{TL-CD $\Diagram[causal]$}
        \label{fig:example-seed:TLCD}
    \end{subfigure}
    \begin{subfigure}[b]{.70\linewidth}
        \centering
        \begin{tikzpicture}[
    scale=.75,
    thick,
    every node/.append style={transform shape},
]
% == NODES =================================================================== %
\node[state,initial,accepting] (u00)
    at (0,-0)
    {$\AutState[causal]_{0}$};
\node[state] (u10)
    at (0,-4)
    {$\AutState[causal]_{1}$};
\node[state,accepting] (u01)
    at (4,-0)
    {$\AutState[causal]_{2}$};
\node[state] (u11)
    at (4,-4)
    {$\AutState[causal]_{3}$};
\node[state] (uX2)
    at (6,-3)
    {$\AutState[causal]_{4}$};

% == EDGES =================================================================== %
\path[->,sloped]

(u00) % -------------------------------------------------------------------- %
edge[loop above] node[sloped=false]
    {$\lnot p  \land  \lnot s$}
    (u00)
edge[bend right=20] node[rotate=180]
    {$p  \land  \lnot s$}
    (u10)
edge[] node[]
    {$\lnot p  \land  s$}
    (u01)
edge[bend right=10] node[shift=(175:0.8)]
    {$p  \land  s$}
    (u11)

(u10) % ----------------------------------------------------------------------- %
edge[out=-100,in=-130,loop] node[sloped=false,shift={(0.5,0)}]
    {$(g \land p)  \land  \lnot s$}
    (u10)
edge[bend right=20] node[swap,shift=(180:0.3)]
    {$(g \land \lnot p)  \land  \lnot s$}
    (u00)
edge[] node[swap]
    {$(g \land p)  \land  s$}
    (u11)
edge[bend right=10] node[swap,shift=(175:1.3)]
    {$(g \land \lnot p)  \land  s$}
    (u01)
edge[bend right=80] node[swap]
    {$\lnot g$}
    (uX2)

(u01) % -------------------------------------------------------------------- %
edge[loop above] node[sloped=false]
    {$\lnot p  \land  \lnot b$}
    (u01)
edge[bend right=20] node[rotate=180]
    {$p  \land  \lnot b$}
    (u11)
edge[bend left] node[]
    {$b$}
    (uX2)

(u11) % ----------------------------------------------------------------------- %
edge[out=-100,in=-130,loop] node[sloped=false,shift={(0.5,0)}]
    {$(g \land p)  \land  \lnot b$}
    (u11)
edge[bend right=20] node[swap]
    {$(g \land \lnot p)  \land  \lnot b$}
    (u01)
edge[bend right] node[]
    {$\lnot g  \lor  b$}
    (uX2)

(uX2) % --------------------------------------------------------------------- %
edge[out=90,in=60,loop] node[sloped=false]
    {$\ltrue$}
    (uX2)

;

\end{tikzpicture}
        \vspace{-15pt}
        \caption{
            DFA $\Autom[causal]$ derived from $\Diagram[causal]$, and equivalent to the LTL\textsubscript{f} formula $\varphi^{\Diagram[causal]} = \lG ( (p) \limplies (\lX g )) \land \allowdisplaybreaks \lG ( (s) \limplies (\lG \lnot \lX b) )$
        }
        \label{fig:example-seed:CausalDFA}
    \end{subfigure}%
    \caption{
        The seed environment.
        The four propositions are $p$ (the agent plants the seed), $g$ (a tree grows), $s$ (the agent sells the seed) and $b$ (the agent buys a tree).
    }
    \label{fig:example-seed}
\end{figure}
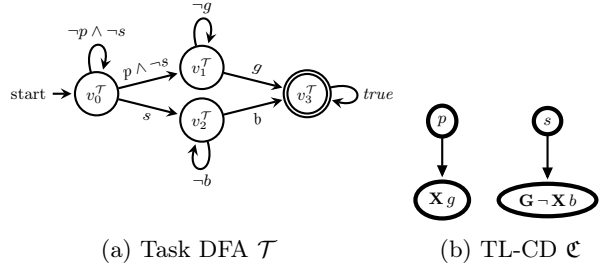
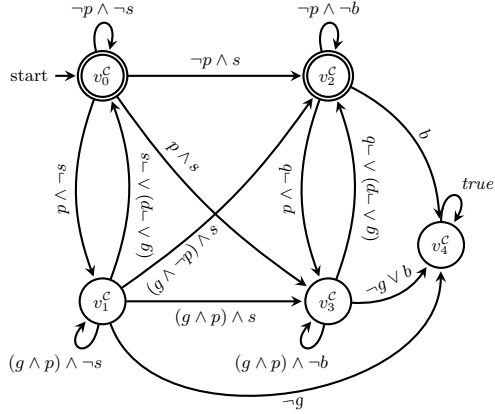

% Let us introduce a running example.
% A farmer has a unique seed, and his goal is to get a tree.
Let us take a running example to illustrate the concept. There is a farmer who possesses a unique seed and his objective is to obtain a tree. 
There are two potential ways to achieve this goal. First, the farmer can plant the seed
($p$) and wait for the tree to grow ($g$).
Alternatively, the farmer can sell the seed (s) and use the money to purchase a tree ($b$). The set of four propositions can thus be represented as $\prop = \{p,g,s,b\}$.
\Cref{fig:example-seed:TaskDFA} illustrates the corresponding task DFA $\Autom[task]$
(note that, because $\Autom[task]$ is deterministic, the transition from $\AutState[task]_{0}$ to $\AutState[task]_{2}$ also fires when $p \land s$ is true, but we assume that the agent cannot take both actions at once).
Additional causal information is provided with the TL-CD $\Diagram[causal]$ (\Cref{fig:example-seed:TLCD}), interpreted as follows:
% $p \causallink \lX \lF g$
% $\lG ( p \limplies \lX \lF g )$
% expresses that planting a tree will result in a tree growing eventually, and
$p \causallink \lX g$ % safety property
expresses that planting a tree will result in a tree growing in the next time-step (e.g., year), and
$s \causallink \lG \lnot \lX b$
% $\lG ( s \limplies \lG \lnot \lX b )$
expresses that selling the seed leads to never being able to buy a tree (as the farmer will never find an offer for a tree that is cheaper than a seed).
% This TL-CD translates into the LTL\textsubscript{f} formula
% % $\lG ( p \limplies \lX \lF g ) \land \allowdisplaybreaks \lG ( s \limplies \lG \lnot \lX b )$,
% $\lG ( p \limplies \lX g ) \land \allowdisplaybreaks \lG ( s \limplies \lG \lnot \lX b )$, % safety property
% which is equivalent to the causal DFA $\Autom[causal]$ (\Cref{fig:example-seed:CausalDFA}).
This TL-CD is equivalent to the \emph{causal} DFA $\Autom[causal]$ illustrated in \Cref{fig:example-seed:CausalDFA} (details are provided later).

\section{Related work}
\textit{Causal inference} answers questions about the mechanism by which manipulating one or a set of variables affects another variable or a set of variables \cite{Spirtes2010}.
In other words, through causal inference, we infer the cause and effect relationships among the variables from  observational data, experimental data, or a combination of both \cite{LeeSanghack2020c}.

Recently, the inherent capabilities of  reinforcement learning (RL) and causal inference (CI) have simultaneously been used for better decision-making including both interventional reasoning \cite{ZhangJunzhe2017,Tennenholtz2019,Lattimore2016,LeeSanghack2019} and counterfactual reasoning \cite{Bareinboim2015,AndrewForney2017a,HowardRonald2005,KOLLER2003} in different settings \cite{LeeSanghack2020b}.
In other words, in an RL setting, harnessing casual knowledge including causal relationships between the actions, rewards, and intrinsic properties of the domain where the agent is deployed can improve the decision-making abilities of the agent \cite{LeeSanghack2020b}.

Usually, incorporating CI in an RL setting can be done using three types of data including observational data, experimental data, and counterfactual data accompanied by the  causal diagram of the RL setting, if available. An agent can have access to observational data by observing another agent, observing the environment, offline learning, acquiring prior knowledge about the underlying setting, etc. Experimental data can be acquired by actively interacting (intervening) with the environment. Counterfactual data can be generated using a specified model, estimated through active learning  empirically \cite{LuChaochao2012,AndrewForney2017a,Pitis2020}. 

In connecting CI and RL, the mentioned data types have been used by themselves or in different combinations.
For example, in \cite{Huang2021}, through sampling observational data in new environments, an agent can make minimal necessary adaptions to optimize the policy given diagrams of structural relationship among the variables of the RL setting.
In \cite{AmyZhang2019}, both observational and experimental data (empirical data) are used to learn \textit{causal states} which are the coarsest partition of the joint history of actions and observations that are maximally predictive of the future in partially observable Markov decision processes (POMDP).
In \cite{AndrewForney2017a}, a combination of all data types has been used in a Multi-Armed Bandit problem in order to improve the personalized decision-making of the agent, where the effect of unmeasured variables (unobserved confounders) has been taken into consideration.

Our research is closely linked to the utilization of formal methods in reinforcement learning (RL), such as RL for reward machines~\cite{IcarteKVM18} and RL with temporal logic specifications~\cite{Aksaray2016,Lixiao2017,Fu2014ProbablyAC,Min2017,Alshiekh2018SafeRL,xu2019joint,AFRAI2021,neider2021advice,zhe_ijcai2019}.
For instance, \cite{IcarteKVM18} proposed a technique known as Q-learning for reward machines (QRM) and demonstrated that QRM can almost certainly converge to an optimal policy in the tabular case.
Additionally, QRM outperforms both Q-learning and hierarchical RL for tasks where the reward functions can be encoded by reward machines.
However, none of these works have incorporated the causal knowledge in expediting the RL process.

% In addition, \textit{temporal causal inference} has been used in learning \cite{LiuYan2010} and reasoning \cite{Ning2018} using time-series data. Learning the temporal structure and causal structure of complex dynamic systems has been studied in \cite{Reiter2022,LiuYan2010}.

\section{Preliminaries}

As typically done in RL problems, we rely on Markov Decision Processes (MDP)~\cite{SuttonB98} to model the effects of sequential decisions of an RL agent.
We, however, deviate slightly from the standard definition of MDPs.
This is to be able to capture temporally extended goals for the agent and thus, want the reward to be non-Markovian.
To capture non-Markovian rewards, we rely on simple finite state machines---deterministic finite automaton (DFA).
Further, to express causal relationships in the environment, we rely on the de facto temporal logic, Linear Temporal Logic (LTL).
%To build reward machines, we label the environment MDP with propositions $\prop$ that indicate relevant high-level knowledge about the environment.
%We assume that the propositions to be known to the RL agent.
We introduce all the necessary concepts formally in this section.
\todo{RR: Running Example to introduce all of the concepts}

\subsubsection{Labeled Markov Decision Process.}
A \emph{labeled Markov decision process}~\cite{IcarteKVM18} is a tuple $\Autom[dyn] = \tuple{\AutStates[dyn], \AutInitState[dyn],  \mdpActions, \AutTransition[dyn], \mdpRewardFunction, \prop, \labelfunc}$ consisting of 
a finite set of states $\AutStates[dyn]$, 
an agent's initial state $\AutInitState[dyn] \in \AutStates[dyn]$, 
a finite set of actions $\mdpActions$,  
a transition probability function $\AutTransition[dyn]: \AutStates[dyn] \times \mdpActions \mapsto \distribution(\AutStates[dyn])$, 
a non-Markovian reward function $\mdpRewardFunction : (\AutStates[dyn] \times \mdpActions)^\ast \times \AutStates[dyn] \mapsto \R$, 
a set of relevant propositions $\prop$, and 
a labeling function $\labelfunc: \AutStates[dyn] \times \mdpActions \times \AutStates[dyn] \mapsto 2^\prop$. 
Here $\distribution(\AutStates[dyn])$ denotes the set of all probability distributions over $\AutStates[dyn]$.
We denote by $\mdpTransition(\AutState[dyn]' | \AutState[dyn], \mdpCommonAction)$ the probability of transitioning to state $\AutState[dyn]'$ from state $\AutState[dyn]$ under action $\mdpCommonAction$.
Additionally, we include a set of propositions $\prop$ that track the relevant information that the agent senses in the environment.
We integrate the propositions in the labeled MDP using the labeling function $\labelfunc$.

% \zhe{Can we use another symbol here for trajectory since pi is already for the policy? RR: Done}
We define a \emph{trajectory} to be the realization of the stochastic process defined by a labeled MDP.
Formally, a trajectory is a sequence of states and actions $\traj = \AutState[dyn]_0 \mdpCommonAction_1 \AutState[dyn]_1 \cdots \mdpCommonAction_k \AutState[dyn]_k $ with $\AutState[dyn]_0 = \AutInitState[dyn]$.
Further, we define the corresponding \emph{label sequence} of $\traj$ as $\traj^L \coloneqq l_0 l_1 l_2 \cdots l_k$ where $l_i = \labelfunc(\AutState[dyn]_i, \mdpCommonAction_{i+1}, \AutState[dyn]_{i+1})$ for each $0\leq i < k$.
%Further, its corresponding \emph{reward sequence} is $r_1 r_2 \cdots r_k$ where $r_i = R(\AutState[dyn]_0 \mdpCommonAction_1 \cdots \mdpCommonAction_i \AutState[dyn]_i)$ for each $0\leq i < k$.

A stationary \emph{policy} $\policy: \AutStates[dyn] \rightarrow \distribution(\mdpActions)$ maps states to probability distributions over the set of actions. In particular, if an agent is in state $\AutState[dyn]_t \in \AutStates[dyn]$ at time step $t$ and is following policy $\policy$, then $\policy(\mdpCommonAction_t | \AutState[dyn]_t)$ denotes its probability of taking action $\mdpCommonAction_t \in \mdpActions$.

% \subsubsection{Reward Machine.} Reward Machines have been proposed recently to express non-Markovian rewards for RL agents~\cite{IcarteKVM18}. 
% They are technically a type of Mealy machine~\cite{shallit-book} that processes label sequences and return a reward sequence.
% \todo{RAJ: Reward machines or DFAs?}

% \begin{definition}%[Reward Mealy machines]
% 	\label{def:rewardMachine}
% 	A \emph{reward machine (RM)} 
% 	$\RM = \langle \RMStates, \RMInitialState, 2^\prop, \RMTransition, \RMOutput \rangle$ consists of 
% 	a finite, nonempty set $\RMStates$ of states, 
% 	an initial state $\RMInitialState \in \RMStates$, 
% 	a finite set $\RMEvents$ of environment events,
% 	a transition function $\RMTransition \colon \RMStates \times 2^\prop \to \RMStates$, 
% 	and an output function $\RMOutput \colon \RMStates \times \RMStates \to \mathbb{R}$.
% \end{definition}

\subsubsection{Deterministic Finite Automaton.}
A \emph{deterministic finite automaton} (DFA) is a finite state machine described using tuple $\Autom = (\AutStates,2^{\prop},\AutTransition,\AutInitState,\AutFinalStates)$
where $\AutStates$ is a finite set of states,
$2^{\prop}$ is the alphabet,
$\AutInitState \in \AutStates$ is the initial state,
$\AutFinalStates \subseteq \AutStates$ is the set of final states, and
$\AutTransition \colon \AutStates \times 2^{\prop} \mapsto \AutStates$ is the deterministic transition function.
We define the size $|\Autom|$ of a DFA as its number of states $|V|$.
%We define the size $\abs{\Autom}$ of a DFA as its number of states $\abs{Q}$.
%

A run of a DFA $\Autom$ on a label sequence $\traj^L = l_0 l_1 \ldots l_{k}\in(2^{\prop})^\ast$, denoted using $\Autom \colon \AutState_0 \xrightarrow{\traj^L} \AutState_{k+1}$,
is simply a sequence of states and labels $\AutState_0 l_0 \AutState_1 l_1 \cdots l_{k} \AutState_{k+1}$, such that $\AutState_0=\AutInitState$ and for each $0 \leq i \leq k$, $\AutState_{i+1} = \AutTransition(\AutState_i,l_i)$.
An accepted run is a run that ends in a final state $\AutState_{k+1}\in \AutFinalStates$.
Finally, we define the language of $\Autom$ as $\lang{\Autom} = \{\traj^L\in (2^{\prop})^\ast~|~\traj^L\text{ is accepted by }\Autom \}$.

We define the parallel composition of two DFAs $\Autom^1 = (\AutStates^1,2^{\prop},\AutTransition^1,\AutInitState^1,\AutFinalStates^1)$ and $\Autom^2 = (\AutStates^2,2^{\prop},\AutTransition^2,\AutInitState^2,\AutFinalStates^2)$ to be the cross-product
$\Autom^1\times\Autom^2 = (\AutStates ,2^{\prop},\AutTransition,\AutInitState,\AutFinalStates^2)$, where $\AutStates = \AutStates^1 \times \AutStates^2$, $\AutTransition((s^1, s^2),  l_i) = (\AutTransition^1(s^1, l_i), \AutTransition^2(s^2, l_i))$, $\AutInitState = (\AutInitState^1, \AutInitState^2)$, and $\AutFinalStates = \AutFinalStates^1 \times \AutFinalStates^2$.
Using such a definition for parallel composition, it is not hard to verify that language $\lang{\Autom^1\times \Autom^2}$ is simply $\lang{\Autom^1}\cap\lang{\Autom^2}$.

\subsubsection{Task DFA.}
Following some recent works~\cite{Memarian00WT20,task-dfa-hmm}, we rely on so-called \emph{task DFA} $\Autom[task]=\tuple{\AutStates[task],2^{\prop},\AutTransition[task],\AutInitState[task],\AutFinalStates[task]}$ to represent the structure of a non-Markovian reward function.
We say a trajectory $\traj$ has a positive reward if and only if the run of $\Autom[task]$ on the label sequence $\traj^L$, $\Autom[task] \colon \AutState[task]_0 \xrightarrow{\traj^L} \AutState[task]_{k+1}$, ends in a final state $\AutState[task]_{k+1}\in \AutFinalStates[task]$.

% \todo{RR: possibly introduce in the next section}

\subsubsection{Linear Temporal Logic.}
\emph{Linear temporal logic} (over finite traces) (LTL\textsubscript{f}) is a logic that expresses temporal properties using temporal modalities.
Formally, we define LTL\textsubscript{f} formulas---denoted by Greek small letters---inductively as:
\begin{itemize}
\item each proposition $p \in \prop$ is an LTL\textsubscript{f} formula; and
\item if $\psi$ and $\varphi$ are LTL\textsubscript{f} formulas, so are $\neg\psi$, $\psi \lor \varphi$, $\lX \psi$ (``neXt''), and $\psi \lU \varphi$ (``Until'').
\end{itemize}
As syntactic sugar, we allow Boolean constants $\ltrue$ and $\lfalse$, and formulas $\psi\land\varphi\coloneqq \neg (\neg \psi \lor \neg \varphi)$ and $\psi\rightarrow \varphi\coloneqq \neg \psi \lor \varphi$.
Moreover, we additionally allow commonly used temporal formulas $\lF\psi\coloneqq \ltrue \lU \psi$ (``finally'') and $\lG\coloneqq \neg \lF\neg \varphi$ (``globally'').
%, which are defined as $\lF\varphi \coloneqq \ltrue\lU\varphi$ and $\lG\varphi \coloneqq \neg\lF(\neg\varphi)$.
%We define $\operators = \{ \lnot, \lor, \land, \limplies, \lX, \lU, \lF, \lG \}~\cup~\Sigma$ to be the set of all operators (which, for simplicity, also incl%We define the size $|\varphi|$ of $\varphi$ as the number of its unique subformulas; e.g., size of $\varphi=(a \lU \lX b) \lor \lX b$ is five, since its five distinct subformulas are 
%$a, b, \lX b, a \lU \lX b$, and $(a \lU \lX b) \lor \lX b$.

To interpret LTL\textsubscript{f} formulas over (finite) trajectories, we follow the semantics proposed by Giacomo and Vardi~\cite{GiacomoV13}.
Given a label sequence $\traj^L$, we define recursively when an LTL\textsubscript{f} formula holds at position $i$, i.e., $\traj^L,i\models \varphi$, as follows: 
\begin{align*}
&\traj^L,i \models p  \text{ if and only if }p \in \traj^L[i]\\
&\traj^L,i \models \lnot \varphi \text{ if and only if } \traj^L,i\not\models \varphi \\
&\traj^L,i \models \varphi \lor \psi \text{ if and only if } \traj^L,i\models \varphi \text{ or }  \traj^L,i\models \psi\\
&\traj^L,i \models \lX \varphi \text{ if and only if } i < |\traj^L| \text{ and } \traj^L,i+1 \models \varphi \\
&\traj^L,i \models \varphi \lU \psi \text{ if and only if } \traj^L,j \models \psi \text{ for some }\nonumber \\
&\hspace{10mm} i \leq j \leq |\traj^L| \text{ and } \traj^L,i' \models \varphi\text{ for all } i \leq i' < j
\end{align*}
We say $\traj^L$ \emph{satisfies} $\varphi$  if $\traj^L\models\varphi$, which, in short, is written as $\traj^L\models \varphi$.

Any LTL\textsubscript{f} formula $\varphi$ can be translated to an equivalent DFA $\Autom^\varphi$, that is, for any $\traj^L\in (2^\prop)^\ast$, $\traj^L\models \varphi$ if and only if $\traj^L\in \lang{\Autom^\varphi}$~\cite{GiacomoV13,ZhuTLPV17}.

\subsubsection{Deterministic causal diagrams.} A causal diagram where every edge represents a cause leading to an effect with probability 1 is called a deterministic causal diagram. In a deterministic causal diagram, the occurrence of the cause will always result in the occurrence of the effect.
\todo{maybe some more context}

\section{Temporal-Logic-Based Causal Diagrams}

We now formalize causality in RL using (deterministic) Causal Diagrams~\cite{GreenlandP11}, a concept that is widely used in the field of Causal Inference.
We here augment Causal Diagrams with temporal logic since we like to express temporally extended relations.
We call such Causal Diagrams as Temporal-Logic-based Causal Diagrams or TL-CDs in short.
While, in principle, TL-CDs can be conceived for several temporal logics, we consider LTL\textsubscript{f} due to its popularity in AI applications~\cite{GiacomoV13}.  
%\begin{figure}
% \includegraphics[scale=0.5]{fig/TLCD-example.png}
%\centering
%\input{fig/TLCD-example.tikz}
%\caption{Example of a Temporal Logic based Causal Diagram (TL-CD)}
%\label{fig:example-tlcd}
%\end{figure}

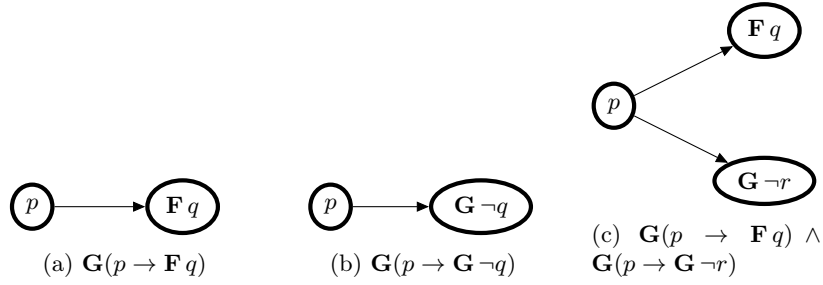
\begin{figure}[t]
\centering
\begin{subfigure}[b]{0.25\textwidth}
% \begin{tikzpicture}[node distance=2cm, every node/.style={draw, circle, minimum size=1cm}]
%     \node[draw] (node1) at (0,0) {$p$};
%     \node[draw] (node2) at (2,0) {$\lF q$};
%     \draw[->,>=triangle 45,line width=0.3] (node1) -- (node2);
% \end{tikzpicture}
\begin{tikzpicture}[node distance=2cm]
    \node[event] (node1) at (0,0) {$p$};
    \node[event] (node2) at (2,0) {$\lF q$};
    \draw[cause] (node1) -- (node2);
\end{tikzpicture}
\caption{$\lG(p\rightarrow \lF q)$}
\label{subfig:tlcd1}
\end{subfigure}
\hspace{0.7cm}
\begin{subfigure}[b]{0.25\textwidth}
\begin{tikzpicture}[node distance=2cm]
    \node[event] (node1) at (0,0) {$p$};
    \node[event] (node2) at (2,0) {$ \lG \neg q$};
    \draw[cause] (node1) -- (node2);
\end{tikzpicture}
\caption{$\lG(p\rightarrow \lG \neg q)$}
\label{subfig:tlcd2}
\end{subfigure}
\hspace{0.5cm}
\begin{subfigure}[b]{0.25\textwidth}
\begin{tikzpicture}[node distance=2cm]
    \node[event] (node1) at (0,0) {$p$};
    \node[event] (node2) at (2,1) {$\lF q$};
    \node[event] (node3) at (2,-1) {$\lG \neg r$};
    \draw[cause] (node1) -- (node2);
    \draw[cause] (node1) -- (node3);
\end{tikzpicture}
\caption{$\lG(p\rightarrow \lF q) \wedge \lG(p\rightarrow \lG \neg r)$}
\label{subfig:tlcd3}
\end{subfigure}
\caption{Examples of TL-CDs with their corresponding description in LTL}
\label{fig:example-tlcd}
\end{figure}

Structurally, for a given set of propositions $\prop$, a \emph{Temporal-Logic-based Causal Diagram} (TL-CD) is a directed acyclic graph $\Diagram[causal]$ where
\begin{itemize}
\item each node represents an LTL\textsubscript{f} formula over propositions $\prop$, and 
\item each edge ($\causallink$) represents a causal link between two nodes.
\end{itemize}

Examples of TL-CDs are illustrated in Figure~\ref{fig:example-tlcd}, where, in the causal relation $\psi \causallink \varphi$, $\psi$ is considered to be the cause and $\varphi$ to be the effect.
The TL-CD in Figure~\ref{subfig:tlcd1} describes that whenever the cause $p$ happens, the effect $q$ eventually (i.e., $\lF q$) occurs.
The TL-CD in Figure~\ref{subfig:tlcd2} describes that whenever the cause $p$ happens, the effect $q$ never (i.e., $\lG \neg q$) occurs.
The TL-CD in Figure~\ref{subfig:tlcd3} describes that whenever the cause $p$ happens, effects $q$ eventually (i.e., $\lF q$) occurs and $r$ never (i.e., $\lG \neg r$) occurs.

For a TL-CD to be practically relevant, we must impose that the occurrence of the cause $\psi$ must precede that of the effect $\varphi$.
To do so, we introduce concepts that track the time of occurrence of an event such as the worst-case satisfaction $w_s(\varphi)$, the worst-case violation $w_v(\varphi)$, the best-case satisfaction $b_s(\varphi)$ and the best-case violation of a formula $\varphi$.
Intuitively, the worst-case satisfaction $w_s(\varphi)$ (resp., the best-case satisfaction $b_s(\varphi)$) tracks the last (resp., the first) possible time point that a formula can get satisfied by a trajectory.
Likewise, the worst-case violation $w_v(\varphi)$ (resp., the best-case violation $b_v(\varphi)$) tracks the last (resp., the first) possible time point that a formula can get violated by a trajectory.
We introduce all the concepts formally in the following definition.
\begin{definition}
For an LTL formula $\varphi$, we define the worst-case satisfaction time $w_{s}(\varphi)$,  
best-case satisfaction time $b_{s}(\varphi)$,  worst-case violation time $w_{v}(\varphi)$ inductively on the structure of $\varphi$ as follows:\\

% \[                                            
% \begin{split}
\begin{align*}
	b_{s}(p)=& w_{s}(p)=b_{v}(p)=w_{v}(p)=0,\\
 	\lnot\varphi:& 
	\begin{cases}
	b_{s}(\lnot\varphi)=b_{v}(\varphi),\\  
	w_{s}(\lnot\varphi)=w_{v}(\varphi),\\
    b_{v}(\lnot\varphi)=b_{s}(\varphi),\\
      w_{v}(\lnot\varphi)=w_{s}(\varphi);
	\end{cases}\\
	\varphi_1\land\varphi_2 :& 
	\begin{cases}
	b_{s}(\varphi_1\land\varphi_2)=\max\{b_{s}(\varphi_1), b_{s}(\varphi_2)\}, \\  
     w_{s}(\varphi_1\land\varphi_2)=\max\{w_{s}(\varphi_1), w_{s}(\varphi_2)\},\\
     b_{v}(\varphi_1\land\varphi_2)=\min\{b_{v}(\varphi_1), b_{v}(\varphi_2)\},\\
     w_{v}(\varphi_1\land\varphi_2)=\max\{w_{v}(\varphi_1), w_{v}(\varphi_2)\};
\end{cases}\\
\varphi_1\lor\varphi_2 :& 
	\begin{cases}
	b_{s}(\varphi_1\lor\varphi_2)=\min\{b_{s}(\varphi_1), b_{s}(\varphi_2)\}, \\  
     w_{s}(\varphi_1\lor\varphi_2)=\max\{w_{s}(\varphi_1), w_{s}(\varphi_2)\},\\
     b_{v}(\varphi_1\lor\varphi_2)=\max\{b_{v}(\varphi_1), b_{v}(\varphi_2)\},\\
     w_{v}(\varphi_1\lor\varphi_2)=\max\{w_{v}(\varphi_1), w_{v}(\varphi_2)\};
\end{cases}\\
\mathbf{G}\varphi:& 
	\begin{cases}
	b_{s}(\mathbf{G}\varphi)=w_{s}(\mathbf{G}\varphi)= w_{v}(\mathbf{G}\varphi)=\infty, \\  
     b_{v}(\mathbf{G}\varphi)=b_{v}(\varphi);
\end{cases}\\
\mathbf{F}\varphi :& 
	\begin{cases}
	b_{s}(\mathbf{F}\varphi)=b_{s}(\varphi), \\  
     w_{s}(\mathbf{F}\varphi)=w_{v}(\mathbf{F}\varphi)= b_{v}(\mathbf{F}\varphi)=\infty;
\end{cases}\\
\mathbf{X}\varphi :& 
	\begin{cases}
	b_{s}(\mathbf{X}\varphi)=b_{s}(\varphi)+1, \\  
     w_{s}(\mathbf{X}\varphi)= w_{s}(\varphi)+1,\\
     w_{v}(\mathbf{X}\varphi)=w_{v}(\varphi)+1,\\
     b_{v}(\mathbf{X}\varphi)=b_{v}(\varphi)+1;
\end{cases}\\
\varphi_1\mathbf{U}\varphi_2 :& 
	\begin{cases}
	b_{s}(\varphi_1\mathbf{U}\varphi_2)=b_{s}(\varphi_2),
 \\  
     w_{s}(\varphi_1\mathbf{U}\varphi_2)=w_{v}(\varphi_1\mathbf{U}\varphi_2)=\infty,\\
     b_{v}(\varphi_1\mathbf{U}\varphi_2)=b_{v}(\varphi_1).
\end{cases}\\
\end{align*}
% \end{split}        \]                             \label{effect}

\end{definition}

For each causal relation $\psi\causallink \varphi$ in a causal diagram $\Diagram[causal]$, we impose the constraints that $w_s(\psi)\leq \min\{b_s(\varphi),b_v(\varphi)\}$ and $w_v(\psi)\leq \min\{b_s(\varphi),b_v(\varphi)\}$.
Such a constraint is designed to make sure that the cause $\psi$, even in the worst-time scenario, occurs before the event $\varphi$, in the best-time scenario.
Based on the constraint, we rule out causal relations in which the cause occurs after the effect such as $\lX p \causallink q$, where $w_s(\lX p)=1$ is greater than $b_s(q)=0$.

To express the meaning of TL-CD in formal logic, we turn to its description in LTL. 
A causal relation $\psi \causallink \varphi$ can be described using the LTL\textsubscript{f} formula $\lG(\psi \rightarrow \varphi)$, which expresses that whenever $\psi$ occurs, $\varphi$ should also occur.
Further, an entire TL-CD $\Diagram[causal]$ can be described using the LTL\textsubscript{f} formula $\varphi^{\Diagram[causal]}\coloneqq \bigwedge_{(\psi\causallink\varphi)} \lG(\psi\rightarrow \varphi)$ which is simply the conjunction of the LTL\textsubscript{f} formulas corresponding to each causal relation in $\Diagram[causal]$.

Based on its description in LTL\textsubscript{f} $\varphi^{\Diagram[causal]}$, we can now define when a trajectory $\pi$ satisfies a TL-CD $\Diagram[causal]$.
Precisely, $\traj$ satisfies $\Diagram[causal]$ if and only if its label sequence $\traj^L$ satisfies $\varphi^{\Diagram[causal]}$.

In the subsequent sections, we also rely on a representation of a TL-CD as a deterministic finite automaton (DFA). 
In particular, for a TL-CD $\Diagram[causal]$, we can construct a DFA $\Autom[causal]^{\Diagram[causal]} = \tuple{\AutStates[causal],2^{\prop},\AutTransition[causal],\AutInitState[causal],\AutFinalStates[causal]}$ from its description in LTL\textsubscript{f} $\varphi^{\Diagram[causal]}$.
We call such a DFA a \emph{causal DFA}.
When the TL-CD is clear from the context, we simply represent a causal DFA as $\Autom[causal]$, dropping its superscript.

In the motivating example (\Cref{sec:motivating-example}), the TL-CD $\Diagram[causal]$ pictured in \Cref{fig:example-seed:TLCD} translates into the LTL\textsubscript{f} formula
% $\varphi^{\Diagram[causal]} = \lG ( p \limplies \lX \lF g ) \land \allowdisplaybreaks \lG ( s \limplies \lG \lnot \lX b )$,
$\varphi^{\Diagram[causal]} = \lG ( p \limplies \lX g ) \land \allowdisplaybreaks \lG ( s \limplies \lG \lnot \lX b )$, % safety property
which is equivalent to the causal DFA $\Autom[causal]$ pictured in \Cref{fig:example-seed:CausalDFA}.

\section{Reinforcement Learning with Causal Diagrams}

We now aim to utilize the information provided in a Temporal-Logic-based Causal Diagram (TL-CD) to enhance the process of reinforcement learning in a non-Markovian setting.
However, in our setting, we assume a TL-CD to be a ground truth about the causal relations in the underlying environment. 
As a result, we must ensure that a TL-CD is compatible with a labeled MDP.

Intuitively, a TL-CD $\Diagram[causal]$ is compatible with a labeled MDP $\mdp$ if all possible trajectories of $\mdp$ respect (i.e., do not violate) the TL-CD $\Diagram[causal]$.
To define compatibility formally, we rely on the cross-product $\overline{\mdp}\times \Autom[causal]$, where $\overline{\mdp}$ is a (non-deterministic) finite state machine representation of $\mdp$ with states $\AutStates[dyn]$, alphabet $2^\prop$, transition $\delta(s,l)=\{s'\in \mdpStates ~|~\labelfunc(s,a,s') = l \text{ for some } a\in\mdpActions\}$, initial state $\AutInitState[dyn]$ and final states $\AutStates[dyn]$, and $\Autom[causal]$ is the causal DFA.

Formally, we say that a TL-CD $\Diagram[causal]$ is \emph{compatible} with an MDP $\mdp$ if from any reachable state $(s,q)\in \overline{\mdp}\times\Autom[causal]$, one can always reach a state $(s',q')\in \overline{\mdp}\times\Autom[causal]$ where $q'$ is a final state in the causal DFA $\Autom[causal]$.
The above formal definition ensures that any trajectory of $\mdp$ can be continued to satisfy the causal relations defined by $\Diagram[causal]$.

We are now ready to state the central problem of the paper.
%\todo{RR: clarify the confusion between finite and infinite LTL (see comment)}
%\todo{JR: What about $\lang{\Autom[dyn] \times \Autom[task]} \subseteq \lang{\Autom[causal]}$, where $\lang{\Autom[dyn] \times \Autom[task]}$ is the set of possible positive trajectories?
%Note that $\lang{\Autom[dyn] \times \Autom[task]} \subseteq \lang{\Autom[dyn]}$.}
% TODO: write recap
\begin{problem}[Non-Markovian Reinforcement learning with Causal Diagrams]
Let $\Autom[dyn]$ be a labeled MDP, $\Autom[task]$ be a task DFA, and $\Diagram[causal]$ be a Temporal Logic based Causal Diagram (TL-CD) such that $\Diagram[causal]$ is compatible with $\Autom[dyn]$. 
Given $\Autom[dyn]$, $\Autom[task]$ and $\Diagram[causal]$, learn a policy that achieves a maximal reward in the environment.
\end{problem}
% \subsection{Q-learning with Causal Reward Machines}

% We now adapt the standard Q-learning algorithm to learn policies with respect to causal reward machines.
% We convert $\Diagram[causal]$ to finite state machines for this algorithm.

% To this end, we convert a causal diagram $\Diagram[causal]$ to an LTL\textsubscript{f} formula $\varphi^{\Diagram[causal]}$.
% We then convert the LTL\textsubscript{f} formula $\varphi^{\Diagram[causal]}$ into a DFA $\Autom[causal]$.

We view TL-CDs as a concise representation of the causal knowledge in an environment.
In our next subsection, we develop an algorithm that exploits this causal knowledge to alleviate the issues of extensive interaction in an online RL setting. 

\subsection{Q-learning with early stopping}
Our RL algorithm is an adaptation of QRM~\cite{IcarteKVM18}, which is a Q-learning algorithm~\cite{SuttonB98} that is typically used when rewards are specified as finite state machines.
On a high level, QRM explores the product space $\mdp \times \Autom[task]$ in many episodes in the search for an optimal policy.
We modify QRM by stopping its exploration early based on the causal knowledge from a TL-CD $\Diagram[causal]$.
Before we describe the algorithm in detail, we must introduce some concepts that aid the early stopping.

For early stopping to work, the learning agent must keep track of whether a trajectory can lead to a reward.
We do this by keeping track of the current configuration in the synchronized run of a trajectory on the product $\Autom[task]\times \Autom[causal]$ of the task DFA and the causal DFA. 
We here identify two particular configurations that can be useful for early stopping: \emph{causally accepting} and \emph{causally rejecting}.
Intuitively, a trajectory reaches a causally accepting configuration if all continuations of the trajectory from the current configuration that satisfy the TL-CD $\Diagram[causal]$ receive a reward of 1 (or a positive reward).
On the other hand, a trajectory reaches a causally rejecting configuration if all continuations of the trajectory from the current configuration that satisfy the TL-CD $\Diagram[causal]$, do not receive a reward. 

We formalize the notion of causally accepting configurations and causal rejecting configurations in the following two definitions. We use the terminology $\Autom_q$ to describe a DFA that is structurally identical to DFA $\Autom$, except that its initial state is $q$.
% If at a certain reward machine state $v$, all the sequences starting from $v$ lead to high rewards or low rewards based on the causal diagram, then the learning agent can already stop at such a state as the current policy can be evaluated without the need of going through the whole episode.

% We want to identify such states $v$ of the DFA converted from the RM $\RM$ such that starting from $v$, for any sequence $\zeta$, if it is accepted by the advice DFA $\mathcal {A}$, then it is accepted by $\RM$, i.e., $L(\advice_{v_{\advice}}) \subseteq L(\machine_v)$, where $L(\advice_{v_{\advice}})$ and $L(\machine_v)$ are the languages starting from $v_{\advice}$ of advice DFA $\mathcal {A}$ and $v$ of the DFA converted from the RM $\RM$, respectively.

% \begin{definition} \label{def:causually-accepting-rejecting}
%     \todo[color=pink]{deprecated def: not on the product $\Autom[task] \times \Autom[causal]$}
%    For a causal reward machine (CRM) $\cmachine=(\RM, \Diagram[causal])$ and the DFA $\mathcal{T}$ converted from $\RM$, a state $v$ of the DFA $\mathcal{T}$ is a causally accepting (rejecting) state if all the sequences starting from $v$ lead to acceptance (rejection) by the DFA $\mathcal{T}$ based on the causal diagram $\Diagram[causal]$.
% \end{definition}

%\bigskip\hrule\bigskip 
\begin{definition}[Causally accepting]
We say $(\AutState[task], \AutState[causal]) \in \AutStates[task] \times \AutStates[causal]$ is causally accepting if for each $\traj^L\in {(2^\prop)}^\ast$ for which the run $\Autom[causal]: \AutState[causal] \xrightarrow{\traj^L} \AutState[causal]_f$ ends in some final state $\AutState[causal]_f \in \AutFinalStates[causal]$, the run $\Autom[task]: \AutState[task] \xrightarrow{\traj^L} \AutState[task]_f$ must also end in some final state $\AutState[task]_f \in \AutFinalStates[task]$. 
Equivalently, we say that $(\AutState[task], \AutState[causal]) \in \AutStates[task] \times \AutStates[causal]$ is causally accepting if  $\lang{\Autom[causal]_{\AutState[causal]}} \subseteq \lang{\Autom[task]_{\AutState[task]}}$.
\end{definition}

\begin{definition}[Causally rejecting]
We say $(\AutState[task], \AutState[causal]) \in \AutStates[task] \times \AutStates[causal]$ is causally rejecting if for each $\traj^L\in {(2^\prop)}^\ast$ for which the run $\Autom[causal]: \AutState[causal] \xrightarrow{\traj^L} \AutState[causal]_f$ ends in some final state $\AutState[causal]_f \in \AutFinalStates[causal]$, the run $\Autom[task]: \AutState[task] \xrightarrow{\traj^L} \AutState[task]_f$ must not end in any final state in $\AutFinalStates[task]$.
Equivalently, we say that $(\AutState[task], \AutState[causal]) \in \AutStates[task] \times \AutStates[causal]$ is causally rejecting if  $\lang{\Autom[causal]_{\AutState[causal]}} \cap \lang{\Autom[task]_{\AutState[task]}}=\emptyset$.
\end{definition}

 \begin{remark}
A configuration $(\AutState[task], \AutState[causal])$ may be neither causally accepting nor causally rejecting.
\end{remark}

To illustrate these concepts, we consider the motivating example introduced in \Cref{sec:motivating-example},
where $\Autom[task]$ and $\Autom[causal]$ are depicted in \Cref{fig:example-seed:TaskDFA,fig:example-seed:CausalDFA}.
The initial state $(\AutState[task]_{0},\AutState[causal]_{0})$ is neither causally accepting nor causally rejecting.
If the agent decides to plant the seed, it encounters a label $p$ and reaches $(\AutState[task]_{1},\AutState[causal]_{1})$, which is causally accepting since the only reachable configurations where $\Autom[causal]$ is accepting $\{(\AutState[task]_{3},\AutState[causal]_{0}), (\AutState[task]_{3},\AutState[causal]_{2})\}$ are accepting for $\Autom[task]$.
If the agent decides to sell the seed instead, it encounters a label $s$ and reaches $(\AutState[task]_{2},\AutState[causal]_{2})$, which is causally rejecting since the only reachable configurations where $\Autom[causal]$ is accepting $\{(\AutState[task]_{2},\AutState[causal]_{2})\}$ are rejecting for $\Autom[task]$.

% There is an alternative way of stating a causally accepting state. 
% To this end, let $\Autom_{\AutState}$ be the DFA which is identical to DFA $\Autom$ except that $\AutState$ is the initial state.

% \begin{definition}
% We say $(\AutState[task], \AutState[causal]) \in \AutStates[task] \times \AutStates[causal]$ is causally accepting if $\lang{\Autom[causal]_{\AutState[causal]}} \subseteq \lang{\Autom[task]_{\AutState[task]}}$.
% We say $(\AutState[task], \AutState[causal])$ is causally rejecting if $\lang{\Autom[causal]_{\AutState[causal]}} \cap \lang{\Autom[task]_{\AutState[task]}} = \emptyset$.
% \end{definition}

\begin{algorithm}
    \caption{causally accepting/rejecting state detection}
    \label{alg:detect-causal-states}
    \DontPrintSemicolon
    \textbf{Input:} Task DFA $\Autom[task]$, Causal DFA $\Autom[causal]$, a pair of states $(\AutState[task], \AutState[causal]) \in \AutStates[task] \times \AutStates[causal]$. \;
    $\AutStates{C}^{\Autom[task]} \gets \emptyset$ \tcp*{set of reachable ``causal states'' of \Autom[task]}
    $\Autom{\mathfrak{P}} \gets \Autom[task] \times \Autom[causal]$ \tcp*{the parallel composition of $\Autom[task]$ and $\Autom[causal]$}
    \ForEach{state $(\AutState[task]_r, \AutState[causal]_r)$ of $\Autom{\mathfrak{P}}$ reachable from $(\AutState[task], \AutState[causal])$}{
        \If{$\AutState[causal]_r \in \AutFinalStates[causal]$}{
            $\AutStates{C}^{\Autom[task]} \gets \AutStates{C}^{\Autom[task]} \cup \{\AutState[task]_r\}$ \;
        }
    }
    \tcp{$(\AutState[task], \AutState[causal])$ is causally accepting if $\AutStates{C}^{\Autom[task]} \subseteq \AutFinalStates[task]$}
    \tcp{$(\AutState[task], \AutState[causal])$ is causally rejecting if $\AutStates{C}^{\Autom[task]} \cap \AutFinalStates[task] = \emptyset$}
    \Return{$\AutStates{C}^{\Autom[task]}$}
\end{algorithm}

We now present the pseudo-code of the algorithm used for detecting the causally accepting and causally rejecting configurations in Algorithm~\ref{alg:detect-causal-states}. 
Intuitively, the algorithm relies on a breadth-first search on the cross-product DFA $\Autom[task]\times \Autom[causal]$ of the task DFA and the causal DFA.
 \begin{remark}
The worst-case runtime of Algorithm~\ref{alg:detect-causal-states} is $\mathcal{O}(|2^{\prop}|\cdot|\Autom[task]|\cdot|\Autom[causal]|)$ since the number of edges in the parallel composition $\Autom[task]\times\Autom[causal]$ can be atmost $|2^{\prop}|\cdot|\Autom[task]|\cdot|\Autom[causal]|$.
\end{remark}

\begin{algorithm}
    \caption{Q-learning with TL-CD}
    \label{alg:cd-qrm}
    \DontPrintSemicolon
    % \State {\textbf{Input:} A labeled MDP $\Autom[dyn]$, a reward machine $\RM$, a causal diagram $\Diagram[causal]$.}
    \textbf{Input:} Labeled MDP $\Autom[dyn]$, Task DFA $\Autom[task]$, Causal diagram $\Diagram[causal]$. \;
    Convert $\Diagram[causal]$ to causal DFA $\Autom[causal]$ \;
    Detect causally accepting and rejecting states of $\Autom[task] \times \Autom[causal]$ \;
    \ForEach{training episode}{
        run QTLCD\_episode
        %Perform QRM, interrupt if causally accepting/rejecting state reached \;
        %\If{the episode was interrupted}{
        %    Update the q-value for the last action with the predicted reward (0 for causally rejecting, 1 for causally accepting) \;
        %}
    }
\end{algorithm}

% \FloatBarrier

%\bigskip\hrule\bigskip

% A deterministic finite automaton (DFA) is a five-tuple $\Autom = \tuple{\AutStates, \Sigma, \AutInitState, \AutTransition, \AutFinalStates}$ where $\AutStates$ is a finite, nonempty set of states, $\Sigma$ is the input alphabet, $\AutInitState \in \AutStates$ is the initial state, $\AutTransition \colon \AutStates \times \Sigma \to \AutStates$ is the transition function, and $\AutFinalStates \subseteq \AutStates$ is the set of final states.
% Runs $\Autom \colon \AutState_1 \xrightarrow{a_1} \AutState_2 \cdots \xrightarrow{a_n} \AutState_{n+1}$ and the language $\lang{\Autom} = \{ u \in \Sigma^\ast \mid \Autom \colon \AutInitState \xrightarrow{u} \AutState, \AutState \in \AutFinalStates\}$ are defined as usual.

% Given a DFA $\Autom = (\AutStates, \Sigma, \AutInitState, \AutTransition, \AutFinalStates)$ and a set $\AutStates{P} \subseteq \AutStates$ of states, we define ${\Autom}[\AutStates{P}] = (\AutStates, \Sigma, \AutInitState, \AutTransition, \AutStates{P})$ (i.e., the DFA $\Autom$ with final states $\AutStates{P}$).

We now expand on our adaptation of the QRM algorithm.
The pseudo-code of the algorithm is sketched in Algorithm~\ref{alg:cd-qrm}.
In the algorithm, we first compute the set of causally accepting or causally rejecting configurations as described in Algorithm~\ref{alg:detect-causal-states}.
Next, like a typical Q-learning algorithm, we perform explorations of the environment in several episodes to estimate the Q-values of the state-action pairs.
However, during an episode, we additionally keep track of the configuration of the product $\Autom[task]\times\Autom[causal]$.
If during the episode, we encounter a causally accepting or causally rejecting configuration, we terminate the episode and update the Q-values accordingly.
\todo{1. Add the exact details to the pseudocode, 2. Add the line number to the explanation paragraph}

\begin{algorithm}
    % \caption{Q-learning with TL-CD episode}
    \caption{QTLCD\_episode}
    \label{alg:cd-qrm-episode}
    \DontPrintSemicolon
    
    \textbf{Hyperparameter:}
    Q-learning parameters,
    episode length $\mathit{eplength}$.
    \;
    
    \textbf{Input:}
    labeled MDP $\Autom[dyn]$,
    task DFA $\Autom[task]$,
    causal DFA $\Autom[causal]$,
    learning rate $\alpha$.
    \;
    
    %\textbf{Input:}
    %a set of q-functions $\setOfQFunctions = \set{\qValue^{\AutState[task]} | \AutState[task] \in \AutStates[task] }$
    %\;
    \textbf{Output:}
    the updated set of q-functions $\setOfQFunctions$
    \;
    
    $\AutState[dyn] \gets \AutInitState[dyn]; \AutState[task] \gets \AutInitState[task]; \AutState[causal] \gets \AutInitState[causal]$
    \tcp*{initialise states}
    
    $R \gets 0$
    \tcp*{initialise cumulative reward}
    
    \For{$0 \leq t < \mathit{eplength}$}{
        
        $\mdpCommonAction \gets \text{GetEpsilonGreedyAction}(\qValue^{\AutState[task]}, \AutState[dyn])$ \label{algLine:choiceOfAction}
        \tcp*{get action from policy}
        
        % $\AutState[dyn]' \gets \mathit{sample}(\AutTransition[dyn](\AutState[dyn],\mdpCommonAction))$
        % \tcp*{execute action in $\Autom[dyn]$}
        % \zhe{ExecuteAction()?}
        $\AutState[dyn]' \gets \mathit{ExecuteAction}(\AutTransition[dyn](\AutState[dyn],\mdpCommonAction))$
        \tcp*{based on distribution $\AutTransition[dyn](\AutState[dyn],\mdpCommonAction)$}

        ${\AutState[task]}' \gets \AutTransition[task](\AutState[task],\labelfunc(\AutState[dyn],\mdpCommonAction,\AutState[dyn]'))$
        \tcp*{synchronize $\Autom[task]$}
        
        ${\AutState[causal]}' \gets \AutTransition[causal](\AutState[causal],\labelfunc(\AutState[dyn],\mdpCommonAction,\AutState[dyn]'))$
        \tcp*{synchronize $\Autom[causal]$}
    
        $R' \gets \mathbb{1}_{\AutFinalStates[task]}({\AutState[task]}')$
        \label{algLine:taskReward}
        % \zhe{?}
        \tcp*{compute cumulative reward}
    
        \tcp{override reward based on causal analysis:}
        \lIf{$({\AutState[task]}',{\AutState[causal]}')$ causally accepting}{$R' \gets 1 $}
        \lIf{$({\AutState[task]}',{\AutState[causal]}')$ causally rejecting}{$R' \gets 0$}
    
        update $\qValue^{\AutState[task]}(\AutState[dyn],\mdpCommonAction)$ using reward $r = R'-R$
        \tcp*{Bellman update}
    
        \lIf{${\AutState[task]}' \in \AutFinalStates[task]$}{\Return \setOfQFunctions}
        \tcp*{end of episode}
        
        \lIf{$({\AutState[task]}',{\AutState[causal]}')$ causally accepting or rejecting}{\Return \setOfQFunctions}
        \tcp*{interrupt episode early}
    
        $\AutState[dyn] \gets \AutState[dyn]'; \AutState[task] \gets {\AutState[task]}'; \AutState[causal] \gets {\AutState[causal]}'; R \gets R'$
    }
    \Return \setOfQFunctions\;
\end{algorithm}

The Q-learning with TL-CD algorithms consists of a loop of several episodes.
The pseudo-code of one episode is sketched in \Cref{alg:cd-qrm-episode}.
The instant reward $r$ is computed such that the cumulative reward $R$ is $1$ if and only if the Task DFA is accepting.
%\Cref{algLine:taskReward}
The cumulative reward is then overridden if it is possible to predict the future cumulative reward, based on if the current configuration is causally accepting or causally rejecting.
If the reward could be predicted, the episode is interrupted right after updating the q-functions, using that predicted reward.
Note that the notion causally accepting and rejecting configurations is defined on unbounded episodes, and might predict a different reward than if the episode were to time out.

\todo[color=magenta]{insert explicit algo for QRM, and mention that this algo is inspired by QRM}

\todo[color=magenta]{add remark about episode length: theory supposes unbounded trajectories, but in practice, we use a timeout}

The above algorithm follows the exact steps of the QRM algorithm and thus, inherits all its advantages, including termination and optimality.
The only notable difference is the early stopping based on the causally accepting and causally rejecting states.
However, when these configurations are reached, based on their definition, all continuations are guaranteed to return positive and no reward, respectively.
Thus, early stopping helps to determine the future reward and update the estimates of Q-value earlier.
We next demonstrate the advantages of the algorithm experimentally.

%\bigskip\hrule\bigskip

% \begin{definition}[Alternative to Definition~\ref{def:causually-accepting-rejecting}]
% \todo[color=pink]{deprecated def: not on the product $\Autom[task] \times \Autom[causal]$}
% Let $\cmachine=(\RM, \Diagram[causal])$ be a causal reward machine,
% $\mathcal A_\RM = \tuple{Q, \Sigma, q_I, \delta, F}$ the DFA obtained from the reward machine $\RM$,
% and $\mathcal A_{\Diagram[causal]}$ the DFA obtained from the causal diagram.
% Then, we call a state $q \in Q$ of $\mathcal A_\RM$ \emph{causally accepting} if $\bigl( L(\mathcal A_\RM[\{ q \}]) \cdot \Sigma^\ast \cap L(\mathcal A_{\Diagram[causal]}) \bigr) \subseteq L(\mathcal A_\RM)$.
% Moreover, we call $q$ \emph{causally rejecting} if $\bigl( L(\mathcal A_\RM[\{ q \}]) \cdot \Sigma^\ast \cap L(\mathcal A_{\Diagram[causal]}) \bigr) \cap L(\mathcal A_\RM) = \emptyset$.
% \end{definition}

% \begin{remark}
%   Note that a state $v$ can be neither a causally accepting state nor a causally rejecting state.  
% \end{remark}

\section{Case Studies}
\label{sec:experiments}

In this section, we implement the proposed 
Q-learning with TL-CD (\method[QTLCD]) algorithm in comparison with a baseline algorithm in three different case studies.

In each case study, we compare the performance of the following two algorithms:
\begin{itemize}
\item Q-learning with TL-CD (\method[QTLCD]): the proposed algorithm, including early stopping of the episodes when a causally accepting/rejecting state is reached
\item Q-learning with Reward Machines (\method[QRM]): the algorithm from \cite{IcarteKVM18}, with the same MDP and RM but no causal diagram.
\todo{explain how to go from Task DFA to Reward Machine}
\end{itemize}

\subsection{Case Study I: Small Office World Domain}
\label{sec:experiments:smalloffice}

\begin{figure}[t]
    \centering
    \begin{subfigure}[b]{0.5\linewidth}
        \centering
        \begin{tikzpicture}[
    scale=0.3,
    every node/.style={scale=2.5},
    every node/.append style={transform shape},
]

\draw[step=1cm,black,very thin] (0,0) grid (17,9);
% Border walls
\fill[black] (0,0) rectangle (1,9);
\fill[black] (2,8) rectangle (11,9);
\fill[black] (12,8) rectangle (17,9);
\fill[black] (0,0) rectangle (17,1);
\fill[black] (16,0) rectangle (17,9);
% Inside the map walls
% Room 1 wall
\draw[black, ultra thick] (3,4) rectangle (5,4);
\draw[black, ultra thick] (5,4) rectangle (5,7);
% Middle walls
\draw[black, ultra thick] (7,0) rectangle (7,4);
\draw[black, ultra thick] (7,5) rectangle (7,8);
\draw[black, ultra thick] (10,0) rectangle (10,4);
\draw[black, ultra thick] (10,5) rectangle (10,8);
\draw[black, ultra thick] (7,4) rectangle (10,4);
\draw[black, ultra thick] (7,5) rectangle (10,5);
% r2 and r3 walls
\draw[black, ultra thick] (12,8) rectangle (12,5);
\draw[black, ultra thick] (12,3) rectangle (12,4);
\draw[black, ultra thick] (12,3) rectangle (13,3);
\draw[black, ultra thick] (13,3) rectangle (13,1);
% Brown walls
%e1
\draw[brown, ultra thick] (1,8) rectangle (2,8);
% at O
\draw[brown, ultra thick] (8,4) rectangle (8,5);
\draw[brown, ultra thick] (9,4) rectangle (9,5);
\draw[brown, very thick] (1,8) rectangle (2,8);
% e2
\draw[brown, ultra thick] (11,8) rectangle (12,8);
% at C
\draw[brown, ultra thick] (12,4) rectangle (12,5);

% Escalator part
\draw[green, ultra thick] (12,4) rectangle (15,4);
\draw[green, ultra thick] (12,5) rectangle (14,5);
\draw[green, ultra thick] (15,4) rectangle (15,7);
\draw[green, ultra thick] (14,5) rectangle (14,6);
\draw[green, ultra thick] (13,7) rectangle (15,7);
\draw[green, ultra thick] (13,6) rectangle (14,6);

% Text labels
\draw[red] (1.5,8.5) node {$e_1$};
\draw[red] (11.5,8.5) node {$e_2$};
\draw (3.5,5.5) node {$k_1$};
\draw (8.5,4.5) node {$O$};
\draw (7.5,4.5) node {$a$};
\draw (9.5,4.5) node {$b$};
\draw (12.5,4.5) node {$c$};
\draw (13.5,6.5) node {$k_2$};

% Arrows
% e1
\draw[ thin,->,green] (1.5,7.5) -- (1.5,8.3);
% e2
\draw[ thin,->,green] (11.5,7.5) -- (11.5,8.3);
% middle
\draw[ thin,->,green] (8.2,4.5) -- (7.7,4.5);
\draw[ thin,->,green] (8.8,4.5) -- (9.3,4.5);
% Escalator arr
\draw[ thin,->,green] (11.7,4.5) -- (12.3,4.5);
\draw[ thin,->,green] (12.7,4.5) -- (13.3,4.5);
\draw[ thin,->,green] (13.7,4.5) -- (14.3,4.5);
\draw[ thin,->,green] (14.5,4.7) -- (14.5,5.3);
\draw[ thin,->,green] (14.5,5.7) -- (14.5,6.3);
\draw[ thin,->,green] (14.3,6.5) -- (13.8,6.5);
\end{tikzpicture}
        \caption{Map of the environment}
        \label{fig:cs-SmallOfficeWorld:Map}
    \end{subfigure}%
    \begin{subfigure}[b]{0.5\linewidth}
        \centering
        \includegraphics[scale=0.22]{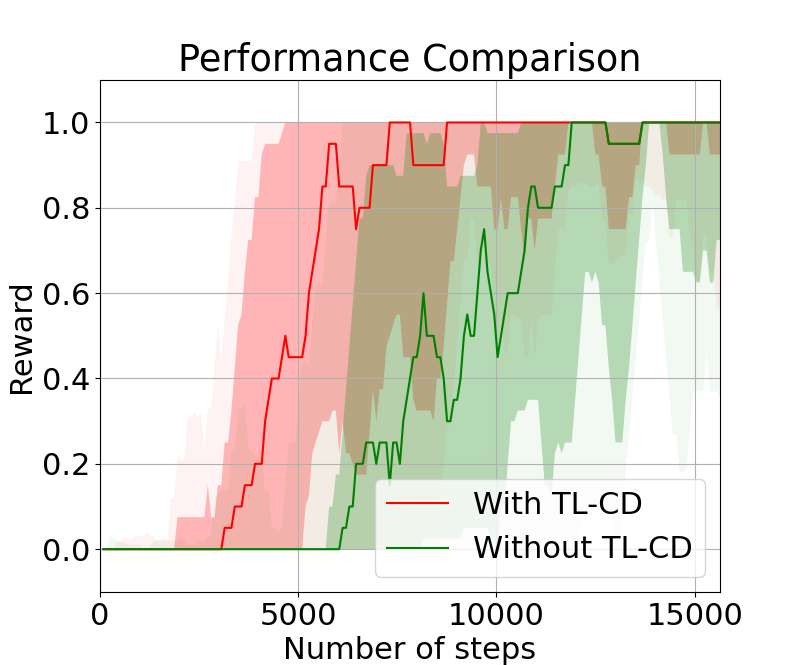}
        \caption{Performance comparison}
        \label{fig:cs-SmallOfficeWorld:results}
        \end{subfigure}
        \\
    \begin{subfigure}[b]{.50\linewidth}
        \centering
        \begin{tikzpicture}[
    scale=.6,
    thick,
    every node/.append style={transform shape},
    every node/.append style={font=\large},
]
% == NODES =================================================================== %
\node[state,initial] (vInit)
    at (0,0)
    {$\AutState[task]_{0}$};
\node[state] (v1)
    at (2,1)
    {$\AutState[task]_{1}$};
\node[state] (v2)
    at (4,1)
    {$\AutState[task]_{2}$};
\node[state] (v3)
    at (2,-1)
    {$\AutState[task]_{3}$};  
\node[state] (v4)
    at (4,-1)
    {$\AutState[task]_{4}$};
\node[state,accepting] (vEnd)
    at (6,0)
    {$\AutState[task]_{5}$};

% == EDGES =================================================================== %
\path[->,sloped]

(vInit) % -------------------------------------------------------------------- %
edge[loop above] node[sloped=false]
    {$\lnot a \land \lnot b$}
    (vInit)
edge[] node[]
    {$a$}
    (v1)
edge[] node[swap]
    {$b$}
    (v3)

(v1) % ----------------------------------------------------------------------- %
edge[loop above] node[sloped=false]
    {$\lnot k_1$}
    (v1)
edge[] node[]
    {$k_1$}
    (v2)

(v2) % ----------------------------------------------------------------------- %
edge[loop above] node[sloped=false]
    {$\lnot e_1$}
    (v2)
edge[above] node[swap]
    {$e_1$}
    (vEnd)

(v3) % ----------------------------------------------------------------------- %
edge[loop below] node[sloped=false]
    {$\lnot k_2$}
    (v3)
edge[] node[swap]
    {$k_2$}
    (v4)

(v4) % ----------------------------------------------------------------------- %
edge[loop below] node[sloped=false]
    {$\lnot e_2$}
    (v4)
edge[] node[swap]
    {$e_2$}
    (vEnd)

(vEnd) % --------------------------------------------------------------------- %
edge[loop above] node[sloped=false]
    {$\ltrue$}
    (vEnd)

;

\end{tikzpicture}
        \caption{Task DFA $\Autom[task]$}
        \label{fig:cs-SmallOfficeWorld:TaskDFA}
    \end{subfigure}%
    \begin{subfigure}[b]{.50\linewidth}
        \centering
        \begin{tikzpicture}[
    scale=.8,
    thick,
    every node/.append style={transform shape},
]
% == NODES =================================================================== %
\node[event] (b)
    at (0,-0)
    {$b$};
\node[event] (c)
    at (2,-0)
    {$c$};
\node[event] (k2)
    at (4,-0.5)
    {$k_2$};

\node[event] (never_e1)
    at (0,-2)
    {$\lG \lnot e_1$};
\node[event] (eventually_k2)
    % at (1,-1)
    % {$\lF k_2$};
    at (2,-1)
    {$\lX \lX \lX \lX \lX k_2$}; % safety property
\node[event] (never_e2)
    at (4,-2)
    {$\lG \lnot e_2$};

% == EDGES =================================================================== %
\path[->,sloped]
(b) edge[cause] node {} (never_e1)
(c) edge[cause] node {} (eventually_k2)
(k2) edge[cause] node {} (never_e2)
;

\end{tikzpicture}
        \caption{TL-CD $\Diagram[causal]$}
        \label{fig:cs-SmallOfficeWorld:TLCD}
    \end{subfigure}%
    \caption{
        Case study I: small office world. The rewards attained in 10 independent simulation runs averaged for every 10 training steps.
        % (\subref{fig:cs-SmallOfficeWorld:Map}) is the map of the small office world.
        % (\subref{fig:cs-SmallOfficeWorld:results}) shows the performance comparison with and without TL-CD over the task given in (\subref{fig:cs-SmallOfficeWorld:TaskDFA}).
        % The TL-CD for small office world is shown in (\subref{fig:cs-SmallOfficeWorld:TLCD}).
    }
    \label{fig:cs-SmallOfficeWorld}
\end{figure}
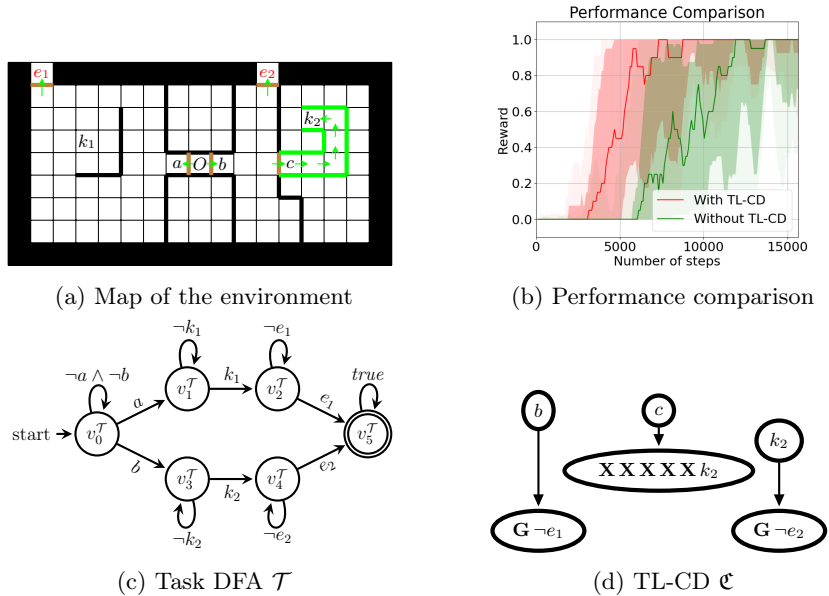

We consider a small officeworld scenario in a 17 $\times$ 9 grid. The agent's objective is to first reach the location of either one key $k_1$ or $k_2$ and then exit the grid by reaching either $e_1$ or $e_2$. The agent navigates on the grid with walls, keys, and one-way doors. The set of actions is $A = \{S, N, E, W\}$. The action $S, N, E, W$ correspond to moving in the four cardinal directions. The one-way doors are shown in \Cref{fig:cs-SmallOfficeWorld:Map} with green arrows. We specify the complex task of the RL agent in a maze as a \textit{deterministic finite automaton (DFA)} $\mathcal{T}$ (see Fig. \ref{fig:cs-SmallOfficeWorld}), as both event sequences $a-k_1-e_1$ (open door $a$, pick up key $k_1$, exit at $e_1$) and $b-k_2-e_2$ (open door $b$, pick up key $k_2$, exit at $e_2$) lead to completion of the task of exiting the maze (receiving reward 1). The agent starts at position $o$. If TL-CD in \Cref{fig:cs-SmallOfficeWorld:TLCD} is true, then the RL agent should never go along the sequence $b-k_2-e_2$ as $k_2\causallink\lG\lnot e_2$ means if the agent picks up key $k_2$ then it can never exit at $e_2$ (as $c$ is a one-way door, so the agent can never get outside Room 3 once it enters $c$ to pick up key $k_2$).

\noindent\textbf{Results:} \Cref{fig:cs-SmallOfficeWorld:results} presents the performance comparison of the RL agent with TL-CD and without TL-CD. It shows that the accumulated reward of the RL agent can converge to its optimal value around 1.5 times faster if the agent knows the TL-CD and learns never to open door $b$.

%We specify the complex task of the RL agent in a maze as a \textit{deterministic finite automaton (DFA)} $\mathcal{T}$ (see Fig. \ref{fig:cs-SmallOfficeWorld}), as both event sequences $a-k_1-e_1$ (open door $a$, pick up key $k_1$, exit at $e_1$) and $b-k_2-e_2$ (open door $b$, pick up key $k_2$, exit at $e_2$) lead to completion of the task of exiting the maze (receiving reward 1).

% \begin{figure}[h]
% 	\centering
% 	\vspace{-3mm}
% 	\includegraphics[width=.99\linewidth]{fig/preliminary.pdf}
% 	\caption{A simple example of completing a task $\mathcal{T}$ with possible TL-CDs $\mathcal{D}_1$ and $\mathcal{D}_2$.}
% 	\label{example_task_confounder}
%  	\vspace{-5mm}
% \end{figure}

%The simulation results in Fig. \ref{fig:cs-SmallOfficeWorld:results} show that the accumulated reward of the RL agent can converge to its optimal value much faster if the agent knows the TL-CD and learns never to open door $b$.
%However, if TL-CD $\mathcal{D}_2$ is true, then opening door $c$ is a \textit{confounder} for causing key $k_2$ to be picked up (as there is a one-way escalator from $c$ to $k_2$) and also causing the agent never to exit at $e_2$, so there is no causal relationship between $k_2$ and $e_2$, and a latent event $u$ could also cause $k_2$ to happen.
%For example, it is possible that another key $k_2$ (not caused by $c$) is in Room 2 near exit $e_2$ and $b-k_2-e_2$ will become the optimal sequence (as it is shorter in distance than $a-k_1-e_1$).
%Therefore, different TL-CDs may significantly change how an RL agent should learn to complete a complex task.

\subsection{Case Study II: Large Office World Domain}
\label{sec:experiments:largeoffice}

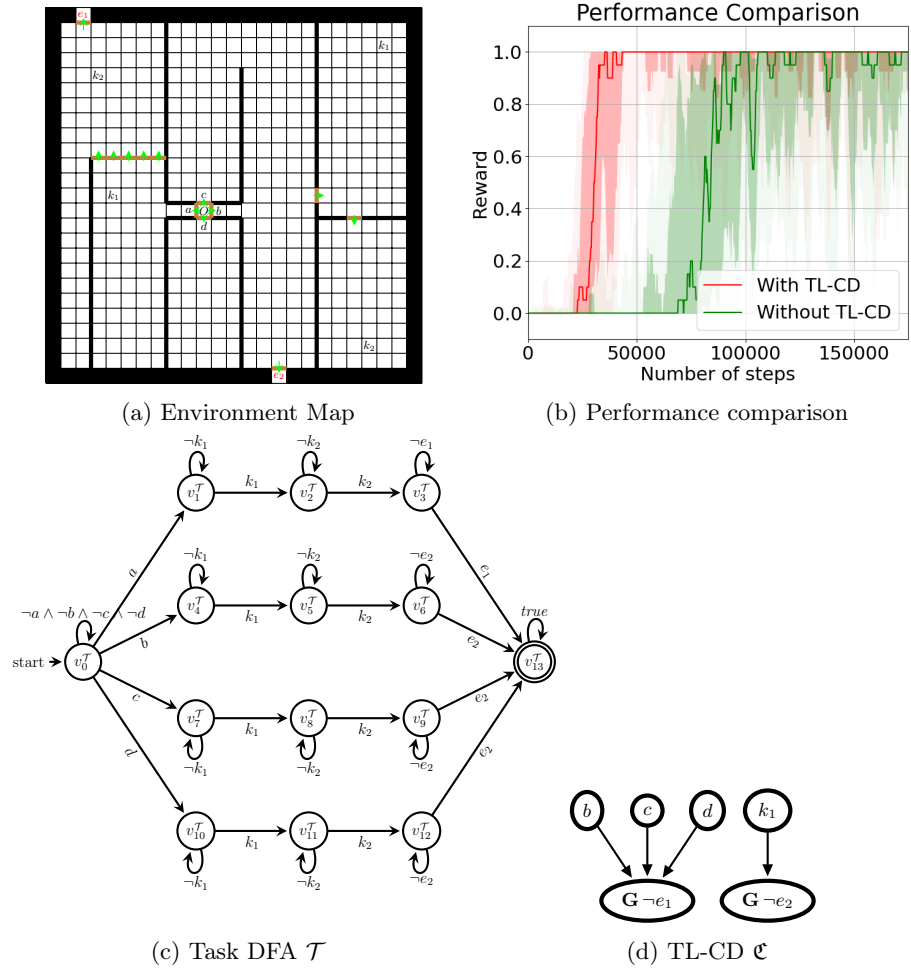
\begin{figure}[t]
    \centering
    \begin{subfigure}[b]{0.5\linewidth}
        \centering
        \begin{tikzpicture}[
    scale=0.2,
    every node/.style={scale=2.5},
    every node/.append style={transform shape},
]

\draw[step=1cm,black,very thin] (0,0) grid (25,25);

% Border walls
\fill[black] (0,0) rectangle (1,25);
\fill[black] (1,24) rectangle (2,25);
\fill[black] (3,24) rectangle (25,25);
\fill[black] (0,0) rectangle (15,1);
\fill[black] (16,0) rectangle (25,1);
\fill[black] (24,0) rectangle (25,25);

% Inside the map walls
% Room 1 wall
\draw[black, ultra thick] (3,1) rectangle (3,15);
% Middle part
\draw[black, ultra thick] (8,24) rectangle (8,12);

\draw[black, ultra thick] (8,1) rectangle (8,11);

\draw[black, ultra thick] (8,11) rectangle (10,11);
\draw[black, ultra thick] (11,11) rectangle (13,11);

\draw[black, ultra thick] (8,12) rectangle (10,12);
\draw[black, ultra thick] (11,12) rectangle (13,12);

\draw[black, ultra thick] (13,21) rectangle (13,12);
\draw[black, ultra thick] (13,11) rectangle (13,1);
% Right Side
\draw[black, ultra thick] (18,24) rectangle (18,13);
\draw[black, ultra thick] (18,12) rectangle (18,1);

\draw[black, ultra thick] (18,11) rectangle (20,11);
\draw[black, ultra thick] (21,11) rectangle (24,11);

% One-way door
% R1
\draw[brown, ultra thick] (3,15) rectangle (8,15);
% Middle part
\draw[brown, ultra thick] (10,12) rectangle (11,12);
\draw[brown, ultra thick] (11,12) rectangle (11,11);
\draw[brown, ultra thick] (11,11) rectangle (10,11);
\draw[brown, ultra thick] (10,11) rectangle (10,12);
%e1
\draw[brown, ultra thick] (2,24) rectangle (3,24);
%e2
\draw[brown, ultra thick] (15,1) rectangle (16,1);
%R2 and R3
\draw[brown, ultra thick] (18,12) rectangle (18,13);
\draw[brown, ultra thick] (20,11) rectangle (21,11);

% Text labels
\draw[red] (2.5,24.5) node {$e_1$};
\draw[red] (15.5,0.5) node {$e_2$};
\draw (10.5,11.5) node {$O$};
\draw (9.5,11.5) node {$a$};
\draw (11.5,11.5) node {$b$};
\draw (10.5,12.5) node {$c$};
\draw (10.5,10.5) node {$d$};
\draw (4.5,12.5) node {$k_1$};
\draw (22.5,22.5) node {$k_1$};
\draw (3.5,20.5) node {$k_2$};
\draw (21.5,2.5) node {$k_2$};

% Arrows
% e1
\draw[thin,->,green] (2.5,23.5) -- (2.5,24.3);
%e2
\draw[thin,->,green] (15.5,1.5) -- (15.5,0.7);
% Middle part
\draw[thin,->,green] (10.2,11.5) -- (9.7,11.5);
\draw[thin,->,green] (10.8,11.5) -- (11.3,11.5);
\draw[thin,->,green] (10.5,11.8) -- (10.5,12.3);
\draw[thin,->,green] (10.5,11.2) -- (10.5,10.7);
% Room 1
\draw[thin,->,green] (3.5,14.8) -- (3.5,15.5);
\draw[thin,->,green] (4.5,14.8) -- (4.5,15.5);
\draw[thin,->,green] (5.5,14.8) -- (5.5,15.5);
\draw[thin,->,green] (6.5,14.8) -- (6.5,15.5);
\draw[thin,->,green] (7.5,14.8) -- (7.5,15.5);
% Room 2 and 3
\draw[thin,->,green] (17.8,12.5) -- (18.5,12.5);
\draw[thin,->,green] (20.5,11.2) -- (20.5,10.5);

\end{tikzpicture}
        \caption{Environment Map}
        \label{fig:cs-LargeOfficeWorld:Map}
    \end{subfigure}%
    \begin{subfigure}[b]{0.5\linewidth}
        \centering
        \includegraphics[scale=0.32]{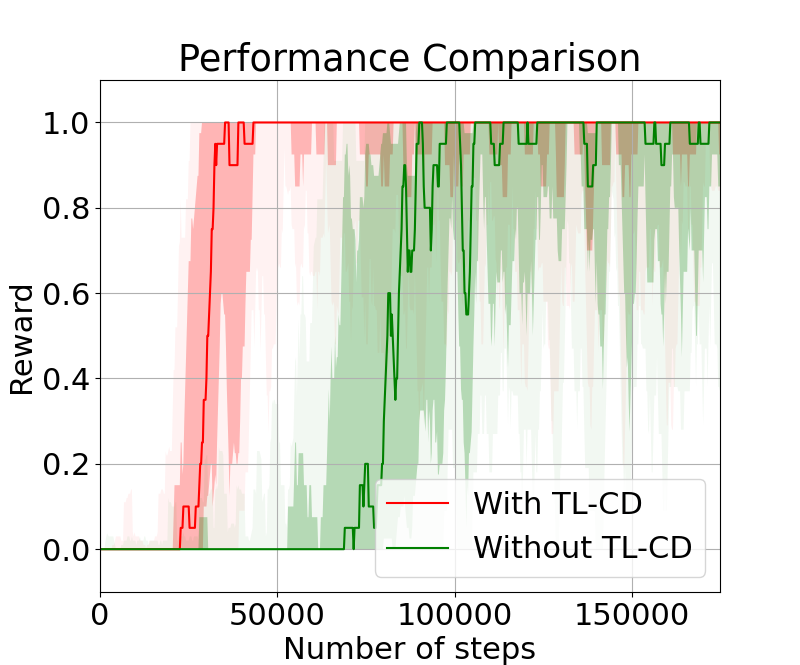}
        \caption{Performance comparison}
        \label{fig:cs-LargeOfficeWorld:results}
        \end{subfigure}
        \\
    \begin{subfigure}[b]{0.5\linewidth}
        \centering
        \begin{tikzpicture}[
    scale=.5,
    thick,
    every node/.append style={transform shape},
    every node/.append style={font=\large},
]
% == NODES =================================================================== %
\node[state,initial] (vInit)
    at (0,0)
    {$\AutState[task]_{0}$};
\node[state] (v1)
    at (3,4.5)
    {$\AutState[task]_{1}$};
\node[state] (v2)
    at (6,4.5)
    {$\AutState[task]_{2}$};
\node[state] (v3)
    at (9,4.5)
    {$\AutState[task]_{3}$};  
\node[state] (v4)
    at (3,1.5)
    {$\AutState[task]_{4}$};
\node[state] (v5)
    at (6,1.5)
    {$\AutState[task]_{5}$};
\node[state] (v6)
    at (9,1.5)
    {$\AutState[task]_{6}$};
\node[state] (v7)
    at (3,-1.5)
    {$\AutState[task]_{7}$};
\node[state] (v8)
    at (6,-1.5)
    {$\AutState[task]_{8}$};
\node[state] (v9)
    at (9,-1.5)
    {$\AutState[task]_{9}$};
\node[state] (v10)
    at (3,-4.5)
    {$\AutState[task]_{10}$};
\node[state] (v11)
    at (6,-4.5)
    {$\AutState[task]_{11}$};
\node[state] (v12)
    at (9,-4.5)
    {$\AutState[task]_{12}$};
\node[state,accepting] (vEnd)
    at (12,0)
    {$\AutState[task]_{13}$};

% == EDGES =================================================================== %
\path[->,sloped]

(vInit) % -------------------------------------------------------------------- %
edge[loop above] node[sloped=false]
    {$\lnot a \land \lnot b \land \lnot c \land \lnot d$}
    (vInit)
edge[] node[]
    {$a$}
    (v1)
edge[] node[swap]
    {$b$}
    (v4)
edge[] node[swap]
    {$c$}
    (v7)
edge[] node[swap]
    {$d$}
    (v10)

(v1) % ----------------------------------------------------------------------- %
edge[loop above] node[sloped=false]
    {$\lnot k_1$}
    (v1)
edge[] node[]
    {$k_1$}
    (v2)

(v2) % ----------------------------------------------------------------------- %
edge[loop above] node[sloped=false]
    {$\lnot k_2$}
    (v2)
edge[above] node[swap]
    {$k_2$}
    (v3)

(v3) % ----------------------------------------------------------------------- %
edge[loop above] node[sloped=false]
    {$\lnot e_1$}
    (v3)
edge[] node[]
    {$e_1$}
    (vEnd)

(v4) % ----------------------------------------------------------------------- %
edge[loop above] node[sloped=false]
    {$\lnot k_1$}
    (v4)
edge[] node[swap]
    {$k_1$}
    (v5)

(v5) % ----------------------------------------------------------------------- %
edge[loop above] node[sloped=false]
    {$\lnot k_2$}
    (v5)
edge[] node[swap]
    {$k_2$}
    (v6)

(v6) % ----------------------------------------------------------------------- %
edge[loop above] node[sloped=false]
    {$\lnot e_2$}
    (v4)
edge[] node[swap]
    {$e_2$}
    (vEnd)

(v7) % ----------------------------------------------------------------------- %
edge[loop below] node[sloped=false]
    {$\lnot k_1$}
    (v7)
edge[] node[swap]
    {$k_1$}
    (v8)

(v8) % ----------------------------------------------------------------------- %
edge[loop below] node[sloped=false]
    {$\lnot k_2$}
    (v8)
edge[] node[swap]
    {$k_2$}
    (v9)

(v9) % ----------------------------------------------------------------------- %
edge[loop below] node[sloped=false]
    {$\lnot e_2$}
    (v9)
edge[] node[swap]
    {$e_2$}
    (vEnd)

(v10) % ----------------------------------------------------------------------- %
edge[loop below] node[sloped=false]
    {$\lnot k_1$}
    (v10)
edge[] node[swap]
    {$k_1$}
    (v11)

(v11) % ----------------------------------------------------------------------- %
edge[loop below] node[sloped=false]
    {$\lnot k_2$}
    (v11)
edge[] node[swap]
    {$k_2$}
    (v12)

(v12) % ----------------------------------------------------------------------- %
edge[loop below] node[sloped=false]
    {$\lnot e_2$}
    (v12)
edge[] node[swap]
    {$e_2$}
    (vEnd)

(vEnd) % --------------------------------------------------------------------- %
edge[loop above] node[sloped=false]
    {$\ltrue$}
    (vEnd)

;

\end{tikzpicture}
        \caption{Task DFA $\Autom[task]$}
        \label{fig:cs-LargeOfficeWorld:TaskDFA}
    \end{subfigure}%
    \begin{subfigure}[b]{0.5\linewidth}
        \centering
        \begin{tikzpicture}[
    scale=.8,
    thick,
    every node/.append style={transform shape},
]
% == NODES =================================================================== %
\node[event] (b)
    at (0,-0)
    {$b$};
\node[event] (c)
    at (1,-0)
    {$c$};
\node[event] (d)
    at (2,-0)
    {$d$};
\node[event] (k1)
    at (3,-0)
    {$k_1$};

\node[event] (never_e1)
    at (1,-1.5)
    {$\lG \lnot e_1$};
\node[event] (never_e2)
    at (3,-1.5)
    {$\lG \lnot e_2$};

% == EDGES =================================================================== %
\path[->,sloped]
(b) edge[cause] node {} (never_e1)
(c) edge[cause] node {} (never_e1)
(d) edge[cause] node {} (never_e1)
(k1) edge[cause] node {} (never_e2)
;

\end{tikzpicture}
        \caption{TL-CD $\Diagram[causal]$}
        \label{fig:cs-LargeOfficeWorld:TLCD}
    \end{subfigure}%
    \label{fig:cs-LargeOfficeWorld}
    \caption{
        Case study II: large office world environment. The rewards attained in 10 independent simulation runs averaged for every 10 training steps.
        % (\subref{fig:cs-LargeOfficeWorld:Map}) is the map of the large office world. (\subref{fig:cs-LargeOfficeWorld:results}) shows the performance comparison with and without TL-CD over the task given in (\subref{fig:cs-LargeOfficeWorld:TaskDFA}). The TL-CD for large office world is shown in (\subref{fig:cs-LargeOfficeWorld:TLCD}).
    }
\end{figure}

%\begin{figure}[ht]
%    \centering
%    \begin{subfigure}[b]{\linewidth}
%        \centering
%        \input{fig/LargeOfficeWorld/map.tikz}
%        \caption{map of the environment}
%        \label{fig:cs-LargeOfficeWorld:TaskDFA}
%    \end{subfigure}%
%    \hfill
%    \begin{subfigure}[b]{0.3\textwidth}
%        \centering
%       \includegraphics[width=1\linewidth]{fig/LargeOfficeWorld/results.png}
%        \caption{Results}
%       \label{fig:cs-LargeOfficeWorld:results}
%    \end{subfigure}
%    \begin{subfigure}[b]{.50\linewidth}
%        \centering
%        \input{fig/LargeOfficeWorld%/TaskDFA.tikz}
%        \caption{Task DFA $\Autom[task]$}
%        \label{fig:cs-LargeOfficeWorld:TaskDFA}
%    \end{subfigure}%
%    \begin{subfigure}[b]{.50\linewidth}
%        \centering
%        \input{fig/LargeOfficeWorld/TLCD.tikz}
%        \caption{TL-CD $\Diagram[causal]$}
%        \label{fig:cs-LargeOfficeWorld:TLCD}
%    \end{subfigure}%
%    \caption{Small Office World}
%    \label{fig:cs-LargeOfficeWorld}
%\end{figure}

We consider a large office world scenario in a $25 \times 25$ grid.
The objective of the agent is to collect both the keys, $k_1$ and $k_2$, and then exit the grid from $e_1$ or $e_2$ (since we assume that there are two locks in both exits, $e_1$ and $e_2$, which needs both keys to unlock). To achieve this objective, the agent needs to collect key $k_1$ first and then key $k_2$ since if it collects key $k_2$ first then it cannot be able to collect key $k_1$ in the map.
The set of actions is $A = \{S, N, E, W\}$. The action $S, N, E, W$ correspond to moving in the four cardinal directions.
The one-way doors are shown in \Cref{fig:cs-LargeOfficeWorld:Map} with green arrows. 
The motivation behind this example is to observe the effect of increasing causally rejecting states on RL agents' performance.
The task DFA and the TL-CD are depicted in \Cref{fig:cs-LargeOfficeWorld}.

\textbf{Results :} \Cref{fig:cs-LargeOfficeWorld:results} presents the performance comparison of the RL agent in a large office world scenario with TL-CD (\method[QTLCD]) and without TL-CD (\method[QRM]).
It shows that the RL agent can converge to its optimal value around 3 times faster if the agent knows the TL-CD.

\subsection{Case Study III: Crossroad Domain}
\label{sec:experiments:crossroad}

\begin{figure}[t]
    \centering
    \begin{subfigure}[b]{0.3\linewidth}
        \centering
        \begin{tikzpicture}[
    scale=.6,
    thick,
    every node/.append style={transform shape},
    every node/.append style={font=\large},
]
% == NODES =================================================================== %
\node[state,initial] (vInit)
    at (0,0)
    {$\AutState[task]_{0}$};
\node[state] (vSink)
    at (2,1)
    {$\AutState[task]_{1}$};
\node[state,accepting] (vGoal)
    at (2,-1)
    {$\AutState[task]_{2}$};

% == EDGES =================================================================== %
\path[->,sloped]

(vInit) % -------------------------------------------------------------------- %
edge[loop above] node[sloped=false]
    {$\lnot c$}
    (vInit)
edge[] node[]
    {$c \land \lnot l$}
    (vSink)
edge[] node[swap]
    {$c \land l$}
    (vGoal)

(vSink) % ----------------------------------------------------------------------- %
edge[loop right] node[sloped=false]
    {$\ltrue$}
    (vSink)

(vGoal) % ----------------------------------------------------------------------- %
edge[loop right] node[sloped=false]
    {$\ltrue$}
    (vGoal)

;

\end{tikzpicture}
        \caption{Task DFA $\Autom[task]$}
        \label{fig:cs-CrossRoad:TaskDFA}
    \end{subfigure}%
    \begin{subfigure}[b]{0.3\linewidth}
        \centering
        \begin{tikzpicture}[
    scale=.8,
    thick,
    every node/.append style={transform shape},
]
% == NODES =================================================================== %
\node[event] (b)
    at (0,-0)
    {$b$};
\node[event] (l)
    at (1.5,-0)
    {$l$};
\node[event] (not_l)
    at (2.5,-0)
    {$\lnot l$};

\node[event] (eventually_l)
    at (0,-1.5)
    {$\lX \lF l$};
    % at (0,-2.2)
    % {$\lX \lX \lX \lX \lX \lX \lX \lX \lX \lX l$}; % safety property
\node[event] (c)
    at (1.5,-1.5)
    {$c$};
\node[event] (not_c)
    at (2.5,-1.5)
    {$\lnot c$};

% == EDGES =================================================================== %
\path[->]
(b) edge[cause] node {} (eventually_l)
% (l) edge[cause,bend right] node {} (c)
% (c) edge[cause,bend right] node {} (l)
(l) edge[cause] node {} (c)
(not_l) edge[cause] node {} (not_c)
;

\end{tikzpicture}
        \caption{TL-CD $\Diagram[causal]$}
        \label{fig:cs-CrossRoad:TLCD}
    \end{subfigure}
    \\
    \vspace{1em}
    \begin{subfigure}[b]{0.6\linewidth}
        \centering
        \includegraphics[scale=0.3]{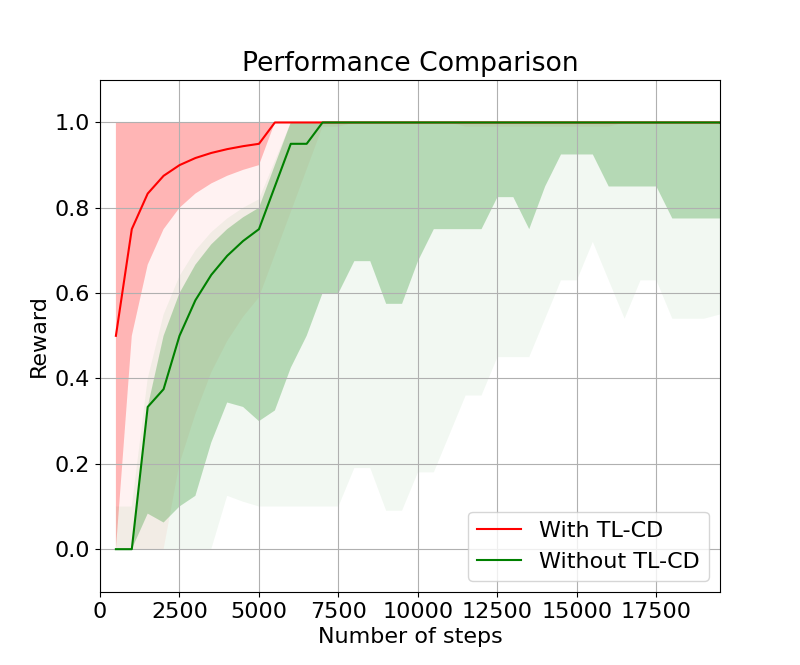}
        \caption{Performance comparison}
        \label{fig:cs-CrossRoad:results}
    \end{subfigure}
    \caption{
        Case study III: crossroad domain. The rewards attained in 10 independent simulation runs averaged for every 10 training steps.
        % The performance comparison of agent in crossroad environment with and without TL-CD is shown in (\subref{fig:cs-CrossRoad:results}). The task DFA is shown in (\subref{fig:cs-CrossRoad:TaskDFA}). The TL-CD for crossroad world is shown in (\subref{fig:cs-CrossRoad:TLCD}).
    }
    \label{fig:cs-CrossRoad}
\end{figure}
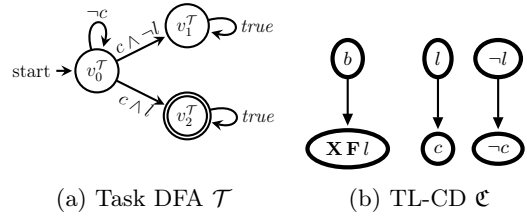
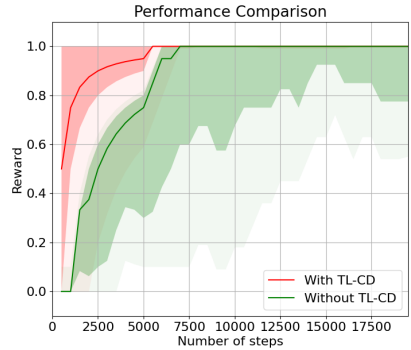

%\begin{figure}[h]
%    \centering
%   \vspace{-3mm}
%    \includegraphics[width=.99\linewidth]{fig/CrossRoad/results.png}
%    \caption{Simulation results of Crossroad domain with and without TLCD}
%    \label{fig:cs-CrossRoad:results}
%    \vspace{-5mm}
%\end{figure}

This experiment is inspired by the real-world example of crossing the road at a traffic signal.
The agent's objective is to reach the other side of the road.
The agent navigates on a grid with walls, crossroad, button, and light signal.
The agent starts from a random location in the grid.
The set of actions is $A = \{S, N, E, W, PressButton, Wait\}$.
The action $PressButton$ presses the button at the crossroad to indicate it wants to cross the road.
After pressing the button, at some later time, the pedestrian crossing light will be turned ON.
 The action $Wait$ will let the agent stay at a location.
The actions $S, N, E, W$ correspond to moving in the four cardinal directions.

To simplify the problem, we make the following assumptions. (1) The agent starts from a fixed location in the left half (one side) of the grid.
(2) The agent knows to cross when the light is ON as after pressing the button the agent has only one valid action. That is, when the light is OFF, it can only wait; and when the light is ON, it will cross.
% After pressing the button at the cross road, the crossing light will be turned on in the next time step.
After pressing the button at the crossroad, the crossing light will turn ON $N$ steps later, where $N$ is a random variable following a geometric distribution of success probability $0.01$.
Thus, the underlying MDP has five observable variables: $x,y$, the discrete coordinates of the agent on the grid,
and $b,p,l$, three Boolean flags that indicate respectively that the button is currently pressed, that the button has been pressed and the light is still OFF, and that the light is turned ON.
% $b$, a Boolean flag that indicates that the button has just been pressed,
% $p$, a Boolean flag that indicates that the button has been pressed and the light is still OFF, and
% $l$, a Boolean flag that indicates the light is turned ON.
% $timer$, the delay timer to add some delay for the light to be turned ON after the button is pressed, we used the timer value of one which means the chance of the light to be turned ON is non-zero, one step after the button is pressed,
% $r$, a Boolean flag that indicates that the button has been pressed and that the light is bound to happen (note that a geometric law is memoryless and does not require extra variables), and
% $l$, a Boolean flag that indicates the light is turned ON.
We specify that task is completed if the agent successfully
% reach the location of the crossroad where the button is located, press the button, and
crosses the road when the light signal is ON (see \Cref{fig:cs-CrossRoad:TaskDFA}).
% We define two labels for the task: $e$ for successfully crossing the road when the light is ON, and $f$ for crossing the road when the light is not ON.

We consider the causal LTL specification $\lG(b \limplies \lF \lX l) \land \lG(l \lequiv c)$ (equivalent to the TL-CD in \Cref{fig:cs-CrossRoad:TLCD}),
where the first part of the conjunction represents the knowledge that the pedestrian light has to turn ON some time later,
and the second part represents the policy of the agent, because we suppose that the agent already knows to cross if and only if the light is on.
Under these conditions, pressing the button leads to a causally accepting state.

\textbf{Results :} \Cref{fig:cs-CrossRoad:results} presents the performance comparison of the RL agent in the crossroad domain with TL-CD (\method[QTLCD]) and without TL-CD (\method[QRM]).
It shows the RL agent will converge faster on average if it knows the TL-CD.
% Although it is unclear if the knowledge of the TL-CD helps converge faster for a median run, convergence seems to be faster and more consistent.

\FloatBarrier

\section{Conclusions and discussions}
This paper introduces the Temporal-Logic-based Causal Diagram (TL-CD) in reinforcement learning (RL) to address the limitations of traditional RL algorithms that use finite state machines to represent temporally extended goals. By capturing the temporal causal relationships between different properties of the environment, our TL-CD-based RL algorithm requires significantly less exploration of the environment and can identify expected rewards early during exploration. Through a series of case studies, we demonstrate the effectiveness of our algorithm in achieving optimal policies with faster convergence than traditional RL algorithms.

In the future, we plan to explore the applicability of TL-CDs in other RL settings, such as continuous control tasks and multi-agent environments. Additionally, we aim to investigate the scalability of TL-CDs in large-scale environments and the impact of noise and uncertainty on the performance of the algorithm. Another direction for future research is to investigate the combination of TL-CDs with other techniques, such as meta-learning and deep reinforcement learning, to further improve the performance of RL algorithms in achieving temporally extended goals.

\noindent\textbf{Acknowledgements.} This work has been supported by NSF CNS 2304863, ONR
N00014-23-1-2505 and DFG grant number 434592664, ONR N00014-22-1-2254, NSF
CNS-1836900

\bibliographystyle{splncs04}
\bibliography{references,references_Xu}

\end{document}